\documentclass{article}

\usepackage{arxiv}

\usepackage[utf8]{inputenc} 
\usepackage[T1]{fontenc}    
\usepackage{hyperref}       
\usepackage{url}            
\usepackage{booktabs}       
\usepackage{amsfonts}       
\usepackage{nicefrac}       
\usepackage{microtype}      
\usepackage{lipsum}		
\usepackage{graphicx}
\usepackage{natbib}
\usepackage{doi}
\usepackage{multirow}
\usepackage{subfigure}
\usepackage{amssymb}
\usepackage{amsmath}
\usepackage{setspace}

\pdfoutput=1
\hypersetup{
    colorlinks=true,
    linkcolor=black,
    citecolor=black,
    filecolor=black,
    urlcolor=black,
}

\title{Semantic Segmentation with Labeling Uncertainty and Class Imbalance}

\date{February, 2021}	

\author{ \href{https://orcid.org/0000-0002-4903-2674}{\includegraphics[scale=0.06]{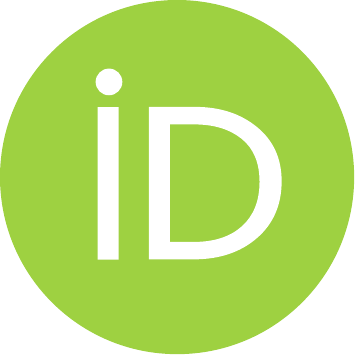}\hspace{1mm}Patrik Ol\~a Bressan} \\
	Federal University of Mato Grosso do Sul\\
	Campo Grande, MS, Brazil \\
	\texttt{patrik.bressan@ifms.edu.br} \\
	\And
	\href{https://orcid.org/0000-0002-9096-6866}{\includegraphics[scale=0.06]{orcid.pdf}\hspace{1mm}José Marcato Junior} \\
	Federal University of Mato Grosso do Sul\\
	Campo Grande, MS, Brazil \\
	\texttt{jose.marcato@ufms.br} \\
	\And
    \href{https://orcid.org/0000-0001-9090-8044}{\includegraphics[scale=0.06]{orcid.pdf}\hspace{1mm}Jos\'e Augusto Correa Martins} \\
	Federal University of Mato Grosso do Sul\\
	Campo Grande, MS, Brazil \\
	\texttt{jose.a@ufms.br} \\
	\And
	\href{https://orcid.org/0000-0002-4527-5724}{\includegraphics[scale=0.06]{orcid.pdf}\hspace{1mm}Diogo Nunes Gon\c{c}alves}\\
	Federal University of Mato Grosso do Sul\\
	Campo Grande, MS, Brazil \\
	\texttt{dnunesgoncalves@gmail.com} \\
	\And
    \href{https://orcid.org/0000-0003-2628-7241}{\includegraphics[scale=0.06]{orcid.pdf}\hspace{1mm}Daniel Matte Freitas} \\
	Federal University of Mato Grosso do Sul\\
	Campo Grande, MS, Brazil \\
	\texttt{daniel.freitas@ufms.br} \\
	\And
    \href{https://orcid.org/0000-0002-0258-536X}{\includegraphics[scale=0.06]{orcid.pdf}\hspace{1mm}Lucas Prado Osco}\\
	University of Western São Paulo\\
	Presidente Prudente, SP, Brazil \\
	\texttt{lucasosco@unoeste.br} \\
	\And
	\href{https://orcid.org/0000-0002-8274-2707}{\includegraphics[scale=0.06]{orcid.pdf}\hspace{1mm}Jonathan de Andrade Silva} \\
	Federal University of Mato Grosso do Sul\\
	Campo Grande, MS, Brazil \\
	\texttt{jonathan.andrade@ufms.br} \\
	\And
	\href{https://orcid.org/0000-0000-0000-0000}{\includegraphics[scale=0.06]{orcid.pdf}\hspace{1mm}Zhipeng Luo} \\
	Xiamen University\\
	Xiamen, FJ, China \\
	\texttt{zpluo@stu.xmu.edu.cn} \\
	\And
	\href{https://orcid.org/0000-0001-7899-0049}{\includegraphics[scale=0.06]{orcid.pdf}\hspace{1mm}Jonathan Li} \\
	University of Waterloo\\
	Waterloo, ON, Canada \\
	\texttt{junli@uwaterloo.ca} \\
	\And
	\href{https://orcid.org/0000-0002-1977-4589}{\includegraphics[scale=0.06]{orcid.pdf}\hspace{1mm}Raymundo Cordero Garcia} \\
	Federal University of Mato Grosso do Sul\\
	Campo Grande, MS, Brazil \\
	\texttt{raymundo.garcia@ufms.br} \\
	\And
	\href{https://orcid.org/0000-0002-8815-6653}{\includegraphics[scale=0.06]{orcid.pdf}\hspace{1mm}Wesley Nunes Gonçalves}\thanks{corresponding author: wesley.goncalves@ufms.br} \\
	Federal University of Mato Grosso do Sul\\
	Campo Grande, MS, Brazil \\
	\texttt{wesley.goncalves@ufms.br} \\
}

\renewcommand{\shorttitle}{\textit{arXiv - Bressan et al. (2021)}}

\hypersetup{
pdftitle={Semantic Segmentation with Labeling Uncertainty and Class Imbalance},
pdfsubject={cs.CV},
pdfauthor={Bressan, et al.},
pdfkeywords={Semantic segmentation, Labeling uncertainty, Class weighting, Loss function},
}

\begin{document}
\maketitle

\begin{abstract}
Recently, methods based on Convolutional Neural Networks (CNN) achieved impressive success in semantic segmentation tasks. However, challenges such as the class imbalance and the uncertainty in the pixel-labeling process are not completely addressed. As such, we present a new approach that calculates a weight for each pixel considering its class and uncertainty during the labeling process. The pixel-wise weights are used during training to increase or decrease the importance of the pixels. Experimental results show that the proposed approach leads to significant improvements in three challenging segmentation tasks in comparison to baseline methods. It was also proved to be more invariant to noise. The approach presented here may be used within a wide range of semantic segmentation methods to improve their robustness.
\end{abstract}

\keywords{Semantic segmentation \and Labeling uncertainty \and Class weighting \and Loss function}

\section{Introduction}
Semantic segmentation is the task of dividing an image into regions whose pixels in the same area have similar properties, whether in color, texture, or belonging to the same object. This task is crucial to infer knowledge of a scene in computer vision systems, as shown in tasks of tree species segmentation~\citep{s20020563}. Recently, significant advances in semantic segmentation have been achieved through Convolutional Neural Networks (CNN), including methods such as SegNet~\citep{segnet_2017}, Fully Convolutional Network (FCN)~\citep{fcn_2015}, and DeepLabv3+~\citep{Chen2018}.

Despite recent advances, two factors have been little explored in the literature during the training of CNNs for semantic segmentation. The first factor is the unbalance of class distribution, where the dominant portions of the data are assigned to a few classes while many classes have little representation in the data. As a consequence, semantic segmentation methods are biased toward the dominant classes during the inference~\citep{LOPEZ2013113}. One way to minimize imbalance is by uniformly sampling data and collecting images (such as well-known image datasets, ImageNet~\citep{Deng2009,ChrabaszczLH17}, MNIST (Modified National Institute of Standards and Technology)~\citep{Lecun98} and CIFAR 10/100), under-sampling the majority classes~\citep{LIU2019105292,TSAI201947,sun2018,Ha2016}, or over-sampling the minority classes \citep{Fernandez2018,Li2017,NEKOOEIMEHR2016405,CASTELLANOS201832}. However, these approaches change the distribution of data and can affect learning and inference \citep{Pozzolo2015}.

The second factor, much less explored in the literature, is related to uncertainty in image labeling~\citep{bulo2017,Bischke2018}. In low resolution or noisy images, the edges of objects become inaccurate and even expert labeling may include annotation errors that impact training. Even in high-resolution images, some objects (e.g., trees~\citep{s20020563}) have complex edges that make them difficult to annotate.

To overcome the aforementioned issues, we propose an approach to deal with class unbalance and uncertainty in the labeling process for image segmentation tasks. For this approach, we introduce a loss function where the contribution of each pixel is weighted. First, pixels belonging to minority classes have their importance increased. Second, pixels near the edges of the object generally have greater uncertainty during labeling and thus have their importance diminished during training. These two pixel-wise weights are combined and produce a great impact during the training and inference of the segmentation methods. We evaluate our approach in three datasets which contain the challenges mentioned above. Experimental results show that the proposed approach provides significant increments when compared to the baseline and state-of-the-art methods.

\section{Related works}

\textbf{Imbalance Data.}
In semantic segmentation, some approaches have been proposed to deal with class imbalance. Traditional approaches use resampling (e.g., oversampling and undersampling) and rebalancing schemes via data statistics, such as inverse or median frequency \citep{Chan2019,Xu2015,BMVC2015_29}. Despite correcting the imbalance, these approaches include several disadvantages. Oversampling methods increase computational cost and may be more prone to overfitting due to the inclusion of duplicate data. On the other hand, undersampling methods can discard important data for learning. Approaches are also based on constraints during training, such as restricting the number of pixels contributing to the loss function during backpropagation at random \citep{BansalCRGR16}, based on the $k$ highest loss of pixels \citep{WuSH16a} or hard samples \citep{Dong2019}. Huang et al. \citep{Huang2016} reduced the effect of class imbalance by enforcing inter-cluster and inter-class margins in standard deep learning frameworks. These margins can be applied through quintuplet instance sampling and the associated triple-header hinge loss. Ren et al. \citep{RenZYU18} proposed a meta-learning framework that assigns weights to training examples based on their gradient directions to reduce class imbalance and corrupted label problems. Recently, focal loss \citep{8417976} was proposed to penalize hard samples assuming that they belong to the minority class. However, this does not happen when minority classes are well defined and may not have their participation in training effectively. A survey on deep learning with class imbalance can be found in \citep{Johnson2019}.

\textbf{Labeling Uncertainty.}
Labeling uncertainty is related to image resolution and object-edge complexity. Similar to this work, Bischke et al. \citep{Bischke2018} applied an adaptive uncertainty weighted class loss to segment satellite imagery. However, only the uncertainty of the class is considered and not the uncertainty of every single pixel, as proposed in this work. Bul\`o et al. \citep{bulo2017} proposed a max-pooling loss that adaptively re-weights the contributions of each pixel based on their observed losses. However, this method does not consider objects whose edges are not well defined and therefore present uncertainties during labeling. Ding et al. \citep{Ding_2019_ICCV} proposed learning boundary objects as an additional class to increase the feature similarity of the same object. Similarly, Shen et al. \citep{Shen2015} addressed the contour detection problem by combining a loss function for contour versus non-contour samples. The labeling uncertainty problem is also related to the size of the object in the image since small objects are harder to label. Islam et al. \citep{islam2017label} proposed a new CNN architecture to predict segmentation labels at several resolutions. A loss function at each stage (scale) provides supervision to improve detail on the segmentation labels. Although it improves the segmentation of object edges, labeling uncertainty is still a problem that degrades the result. Hamaguchi et al. \citep{HamaguchiFNIH18} proposed a novel architecture called local feature extraction which aggregates local features with decreasing dilation factor to segment small objects in remote sensing imagery.

\section{Improving Semantic Segmentation with Labeling Uncertainty and Class Imbalance}

The purpose of semantic segmentation methods is to assign a label to each pixel $x$ of an image $I(x)$, providing a pixel-level mask $\hat{M}(x)$. The most common methods for this task are based on CNNs composed of convolution, pooling, and upsampling layers \citep{fcn_2015,segnet_2017}. This way, the pixel-level mask $\hat{M}$ is obtained through a CNN $f_{\theta}$ with layer parameters $\theta$, $\hat{M} = f_{\theta}(I)$. The dominant loss function used to train a CNN takes the following form:

\begin{equation}
    \min_{\theta \in \Theta} \sum_{(I,M) \in T} L(\hat{M}, M) + \lambda R(\theta)
\end{equation}
where $(I,M)$ is an example consisting of an image $I$ and a ground-truth mask $M$ of the training set $T$, $\hat{M} = f_{\theta}(I)$ is the predicted mask, $L$ is a loss function (e.g., cross-entropy) that penalizes the wrong labels, and $R$ is a regularizer.

In semantic segmentation tasks, the loss function $L$ is usually decomposed into a sum of pixel losses according to Eq. \ref{eq:pixel-loss}. The weight of each pixel contributes uniformly during training.

\begin{equation}
    L(\hat{M}, M) = \frac{1}{n} \sum_{x=1}^n L(\hat{M}(x), M(x))
    \label{eq:pixel-loss}
\end{equation}
where $n$ is the number of pixels.

There are two main issues within this approach: i) class imbalance; and ii) uncertainty in the annotation. The consequence of class imbalance is a bias towards the dominant classes over the classes that occupy smaller parts in the image. This occurs in most real-world image segmentation problems, where few classes dominate most images. Also, some classes do not have well-defined borders (e.g., trees), resulting in uncertainly labeled pixels. An incorrectly labeled pixel influences model learning, making filter convergence and learning even more difficult for small objects.

Figs. \ref{fig:tree} and \ref{fig:rea} present examples that illustrate the challenges of semantic segmentation methods. The trees in Fig. \ref{fig:tree} show that in most images the foreground covers fewer pixels than the background (class imbalance). Besides, trees have edges that are difficult to label, and some pixels may be incorrectly labeled. Fig. \ref{fig:rea} also illustrates the labeling challenge, in which some parts of the object are not visible in the image due to noise when capturing images. 

\subsection{Proposed Loss Function}
To improve these issues, we propose to weight the contribution of each pixel based on its labeled class importance and uncertainty of its labeling.
A weight for each pixel $w(x)$ is used in the loss function according to Eq. \ref{eq:new-pixel-loss}.

\begin{equation}
    L(\hat{M}, M) = \frac{1}{n} \sum_{x=1}^n \omega(x) \cdot L(\hat{M}(x), M(x))
    \label{eq:new-pixel-loss}
\end{equation}

Unlike other approaches (e.g., focal loss \citep{8417976}), the weight $\omega(x)$ of the pixel $x$ is calculated by considering two important characteristics as shown in Eq. \ref{eq:weight}. The first part $\varphi^{c(x)}$ considers class unbalance, where $c(x)$ is the class labeled for pixel $x$. The second part $\delta(x)$ considers the labeling uncertainty of the pixel $x$.
Both parts are described in detail in the sections below.

\begin{equation}
    \omega(x) = \varphi^{c(x)} \cdot \delta(x)
    \label{eq:weight}
\end{equation}

\subsection{Dealing with Class Imbalance}
The first characteristic takes into account the unbalance of classes. To determine the weight of each class $c$, we use the training set according to Eq. \ref{eq:weight-class}. The lower the number of pixels in a given class, the higher the weight so that CNN layer filters fit evenly. When $\omega^c$ equals 1 for all classes, training is performed as traditionally. It is important to note that this weight is the same for all pixels in the same class.

\begin{equation}
\omega^c = \frac{m}{C*n^c}
\label{eq:weight-class}
\end{equation}
where $m$ is the number of pixels of all training images, $C$ is the number of classes, and $n^c$ is the number of pixels that belong to class $c$.

\subsection{Dealing with Labeling Uncertainty}
The second characteristic considers the labeling uncertainty and is calculated for each pixel in the image. This is especially true for objects with poorly defined edges or low-resolution images.
We consider that the closer to the edge, the greater the uncertainty of the class labeled for a given pixel. On the other hand, pixels near the center of objects are labeled more accurately.
This feature can be modeled by Eq. \ref{eq:weight-distance} considering the distance of a pixel to the edges. The main parameter $\sigma$ determines the spread of uncertainty around the edge.

\begin{equation}
    \delta(x) = 1 - e^{-\frac{d(x)^2}{2 \sigma^2}}
    \label{eq:weight-distance}
\end{equation}
where $d(x)$ is the distance from the pixel $x$ to the nearest edge pixel (can be calculated efficiently using the Euclidean Distance Transform) and $\sigma$ is the standard deviation.

\begin{figure}[ht]
	\centering
	\includegraphics[width=.9\columnwidth]{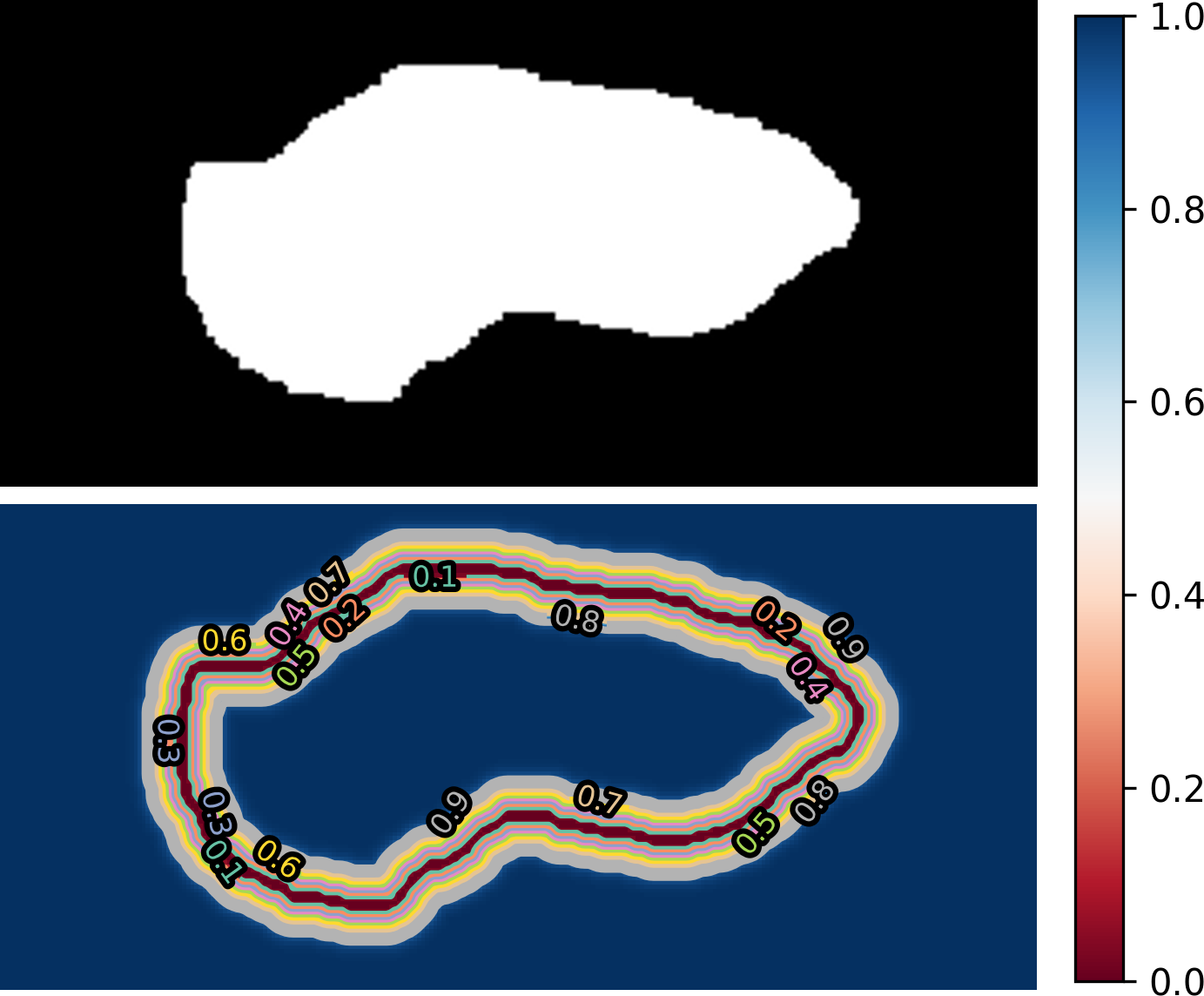}
	\caption{Example of calculating the uncertainty $\delta(x)$ of each pixel $x$. As a pixel approaches the edge, the greater its uncertainty.}
	\label{fig:edt}
\end{figure}

Figure \ref{fig:edt} illustrates the process of calculating $\delta(x)$ for each pixel $x$.
It is possible to observe that the closer to the object's edge, the lower the value of $\delta(x)$ and therefore it is considered as a pixel with high uncertainty.
As a given pixel moves away from the edge, its uncertainty in the labeling is reduced.

\subsection{Semantic Segmentation Methods}
To evaluate the proposed approach, we used two well-known semantic segmentation methods: SegNet~\citep{segnet_2017} and FCN~\citep{fcn_2015}. SegNet~\citep{segnet_2017} is a CNN with encoder and decoder networks, with a final pixel-wise classification layer. For each input, the encoder provides a low-resolution activation map representing the most important features. In this work, the encoder is composed of the convolutional and max-pooling layers of VGG16~\citep{Simonyan14c}. Then, the segmented image is reconstructed by the decoder. The decoder network is composed of convolutional and upsampling layers that use the corresponding max-pooling indices from the encoder to upsample the low-resolution feature map. In the last layer, a softmax classifier receives the feature map from the decoder for pixel-wise classification.

FCN~\citep{fcn_2015} extended classification CNN (VGG16~\citep{Simonyan14c}) by transforming it into fully convolutional, where the fully connected layers were replaced by convolutional layers. In this way, the first part produces a feature map with low-resolution from the image, which is upsampled to produce pixel-wise predictions for segmentation.

\section{Experiments and Results}

\subsection{Image Datasets}
Three image datasets were used in this study to demonstrate the robustness of our method.
These datasets, described below, have the challenges of class imbalance and labeling uncertainty.

\begin{figure}[ht]
	\centering
	\subfigure{\includegraphics[width=.45\columnwidth]{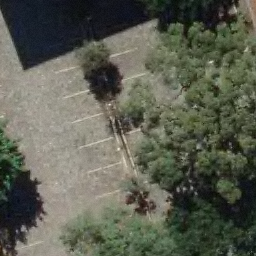}}
	\subfigure{\includegraphics[width=.45\columnwidth]{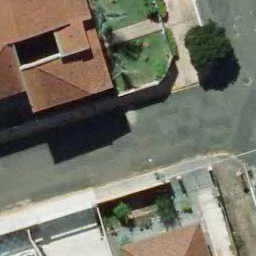}}
	\subfigure{\includegraphics[width=.45\columnwidth]{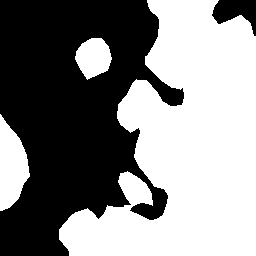}}
	\subfigure{\includegraphics[width=.45\columnwidth]{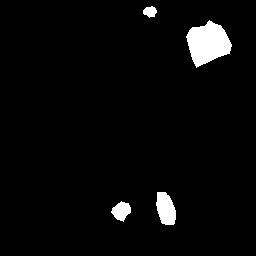}}
	\caption{Sample images from Urban Tree (UT) dataset.}
	\label{fig:tree}
\end{figure}

\textbf{Urban Tree (UT).} 
This dataset is composed of aerial RGB orthoimages generated with a GSD (Ground Sample Distance) of 10 cm from Campo Grande municipality in Brazil. Examples of Urban Tree dataset in Fig. \ref{fig:tree} show that the boundaries of the trees are difficult to label. This dataset is composed by 966 non-overlapping patches of $256 \times 256$ pixels. In the experiments, 580, 193 and 193 patches were used for training, validation, and testing, respectively.

\begin{figure}[ht]
	\centering
	\subfigure{\includegraphics[width=.45\columnwidth]{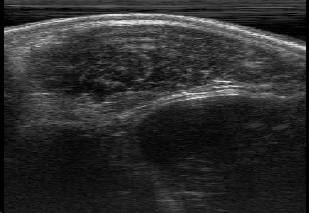}}
	\subfigure{\includegraphics[width=.45\columnwidth]{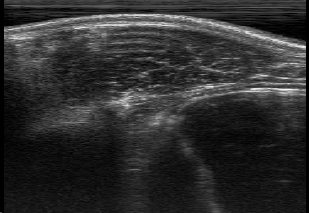}}
	\subfigure{\includegraphics[width=.45\columnwidth]{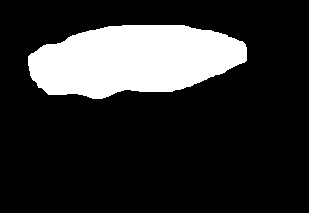}}
	\subfigure{\includegraphics[width=.45\columnwidth]{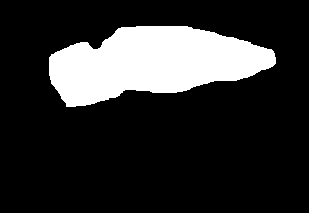}}
	\caption{Sample images from Rib Eye Area (REA) dataset.}
	\label{fig:rea}
\end{figure}

\textbf{Rib Eye Area (REA).}
This image dataset consists of ultrasound images of the Longissimus dorsi muscle between the 11th and 13th ribs of cattle. The goal is to automatically calculate the rib eye area (REA), an important region for decision making during cattle breeding. The main challenge is the uncertainty in the REA annotation, since the image is noisy and even experts have difficulty in delimiting the borders of this region. Fig. \ref{fig:rea} presents examples of images and the annotation made by a specialist. We can observe that some borders are absent and depend on the subjectivity and knowledge of the annotator. To evaluate the segmentation methods, 76 images with $309 \times 213$ resolution were obtained and labeled by an expert. Due to the number of images, the division of the images in training and testing followed 5-fold cross-validation.

\begin{figure}[ht]
	\centering
	\subfigure{\includegraphics[width=.45\columnwidth]{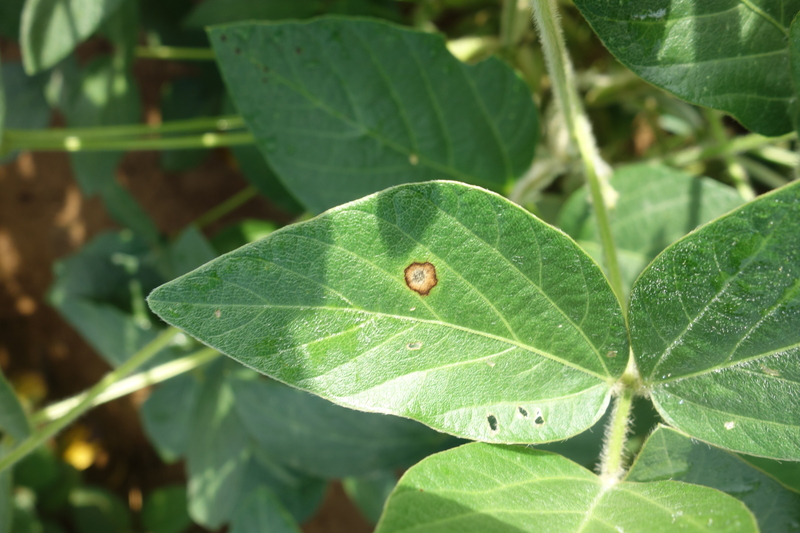}}
	\subfigure{\includegraphics[width=.45\columnwidth]{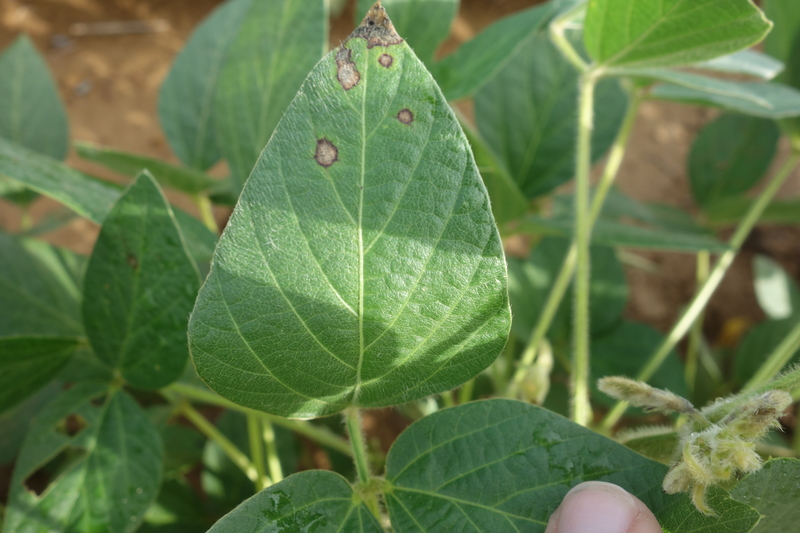}}
	
	\subfigure{\includegraphics[width=.45\columnwidth]{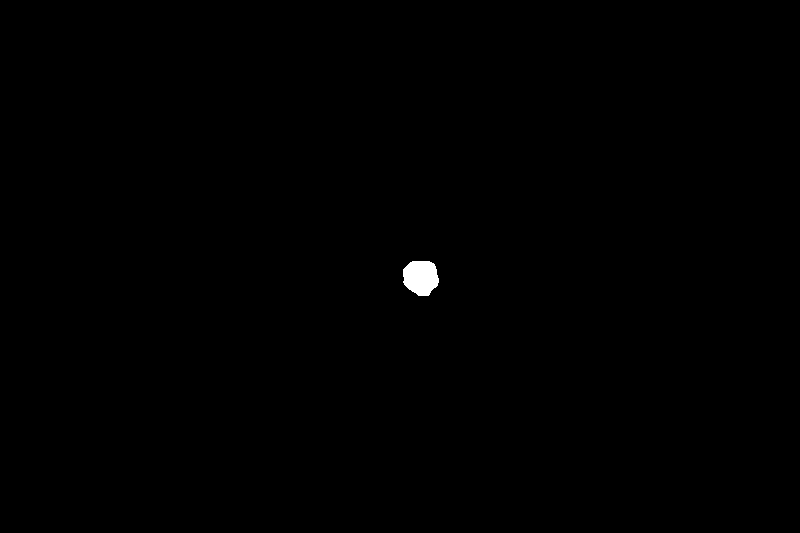}}
	\subfigure{\includegraphics[width=.45\columnwidth]{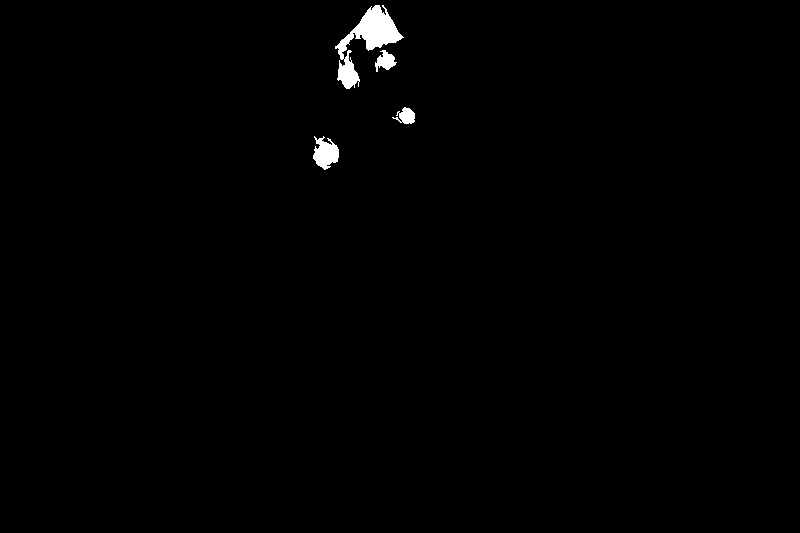}}

	\caption{Sample images from Soybean Disease (SD) dataset.}
	\label{fig:soybean_disease}
\end{figure}

\textbf{Soybean Disease (SD).}
The images from this dataset were obtained through PlantVillage \citep{HughesS15}, which contains several photographs taken by a cell phone in soybean plantations. To compose the image dataset, 201 images with frog-eye disease were identified and manually annotated as shown in Fig. \ref{fig:soybean_disease}. It is important to emphasize that the images were taken in the field, and present several lighting challenges. The images were randomly divided into three sets: 121 for training, 40 for validation, and 40 for testing.

\subsection{Experimental Setup}

For REA and Urban Tree datasets, the images have been resized to $256 \times 256$ pixels. For Soybean Disease dataset, we resized the images to $1024 \times 1024$ pixels because the original ones have high resolution. For all segmentation methods, we use Stochastic Gradient Descent (SGD) optimizer with learning rate of 0.001, momentum of 0.9, and weight decay of 0.0005. The backbone weights of the segmentation methods started with pre-trained weights on ImageNet.

To evaluate the proposed approach and baselines, we use the following popular segmentation metrics: pixel accuracy (PA) and intersection over union (IoU). PA is the percentage of pixels correctly classified for each class. On the other hand, IoU is given by dividing the intersection area by the union area between prediction and ground-truth. Since the background is dominant in most images, we report the PA and IoU results only for the class of interest (e.g., REA, tree and diseases).

\subsection{Results}

In Tables \ref{tab:results_segnet} and \ref{tab:results_fcn}, we compare the baseline methods and the proposed approach using SegNet and FCN, respectively. The main parameter of the proposed approach is $\sigma$ that corresponds to the spread of uncertainty used in the loss function.
Therefore, results for different values of $\sigma$ were also reported.

For SegNet (Table \ref{tab:results_segnet}), the proposed approach improved pixel accuracy (e.g., from 0.744 to 0.838 in Urban Tree dataset, 0.888 to 0.927 in REA dataset, and 0.35 to 0.777 in Soybean Disease dataset). The proposed approach also showed superior IoU results, especially in Urban Tree and SD datasets, where IoU improved from 0.676 to 0.705, and from 0.324 to 0.567, respectively. Further, it is found that using $\sigma = 2$ provided the best result in Urban Tree and SD datasets, while $\sigma = 3$ provided the best one in REA dataset. A lower value of $\sigma$ for Urban Tree and SD datasets is expected due to the size of the foreground (tree and disease), which generally occupies a smaller area than the background compared to REA.

\begin{table}[ht]
\begin{center}
\caption{Comparative results between the proposed approach using SegNet and baseline in the three image datasets.}
\label{tab:results_segnet}
\begin{tabular}{|c|c|c||c|c||c|c|}
\hline
\multicolumn{1}{|c|}{\multirow{2}{*}{\textbf{Method}}} & \multicolumn{2}{c||}{\textbf{Urban Tree}} & \multicolumn{2}{c||}{\textbf{REA}} & \multicolumn{2}{c|}{\textbf{SD}} \\ \cline{2-7} 
\multicolumn{1}{|c|}{}                        & \textbf{PA}          & \textbf{IoU}         & \textbf{PA}           & \textbf{IoU} & \textbf{PA}           & \textbf{IoU}          \\ \hline
SegNet  & 0.744 & 0.676 & 0.888 & 0.841 & 0.350 & 0.324 \\ \hline
SegNet + $\sigma = 1$  & 0.812	& 0.700 & 0.918 & 0.852 & 0.687 & 0.510\\ \hline
SegNet + $\sigma = 2$  & \textbf{0.838}	& \textbf{0.705} & 0.918 & 0.852 & \textbf{0.777} & \textbf{0.567}\\ \hline
SegNet + $\sigma = 3$  & 0.805	& 0.698 & \textbf{0.927} & \textbf{0.853} & 0.668 & 0.509 \\ \hline
\end{tabular}
\end{center}
\end{table}

The proposed approach also provided better results using the FCN. From Table \ref{tab:results_fcn} it is observed that the results increase with the inclusion of the proposed approach. In Urban Tree, REA, and SD datasets, considerable increases of 8\%, 0.5\% and 23.9\% were obtained in the pixel accuracy. On the other hand, IoU obtained by the proposed approach was slightly higher in Urban Tree and REA datasets and lower in SD dataset. Hence, the approach described here has proven to be effective for three datasets that include the challenges of class imbalance and labelling uncertainty and for two semantic segmentation methods.

\begin{table}[ht]
\begin{center}
\caption{Comparative results between the proposed approach using FCN and baseline in the three image datasets.}
\label{tab:results_fcn}
\begin{tabular}{|c|c|c||c|c||c|c|}
\hline
\multicolumn{1}{|c|}{\multirow{2}{*}{\textbf{Method}}} & \multicolumn{2}{c||}{\textbf{Urban Tree}} & \multicolumn{2}{c||}{\textbf{REA}} & \multicolumn{2}{c|}{\textbf{SD}} \\ \cline{2-7} 
\multicolumn{1}{|c|}{}                        & \textbf{PA}          & \textbf{IoU}         & \textbf{PA}           & \textbf{IoU} & \textbf{PA}           & \textbf{IoU}          \\ \hline
FCN  & 0.820 & 0.730 & 0.966 & 0.861 & 0.750 & \textbf{0.611} \\ \hline
FCN + $\sigma = 1$  & 0.892	& 0.754 & 0.967 & \textbf{0.866} & \textbf{0.989} & 0.368 \\ \hline
FCN + $\sigma = 2$  & \textbf{0.900}	& \textbf{0.760} & 0.967 & 0.863 & \textbf{0.989} & 0.371 \\ \hline
FCN + $\sigma = 3$  & 0.896	& 0.729 & \textbf{0.971} & 0.865 & 0.982 & 0.425\\ \hline
\end{tabular}
\end{center}
\end{table}

\subsection{Discussion and Qualitative Results}

As shown in the previous section, FCN achieved better results than SegNet in the three image datasets. Therefore we discuss and present visual results of the FCN baseline and FCN using the proposed approach.

\begin{figure}[!ht]
	\centering
	\subfigure{\includegraphics[width=.32\columnwidth]{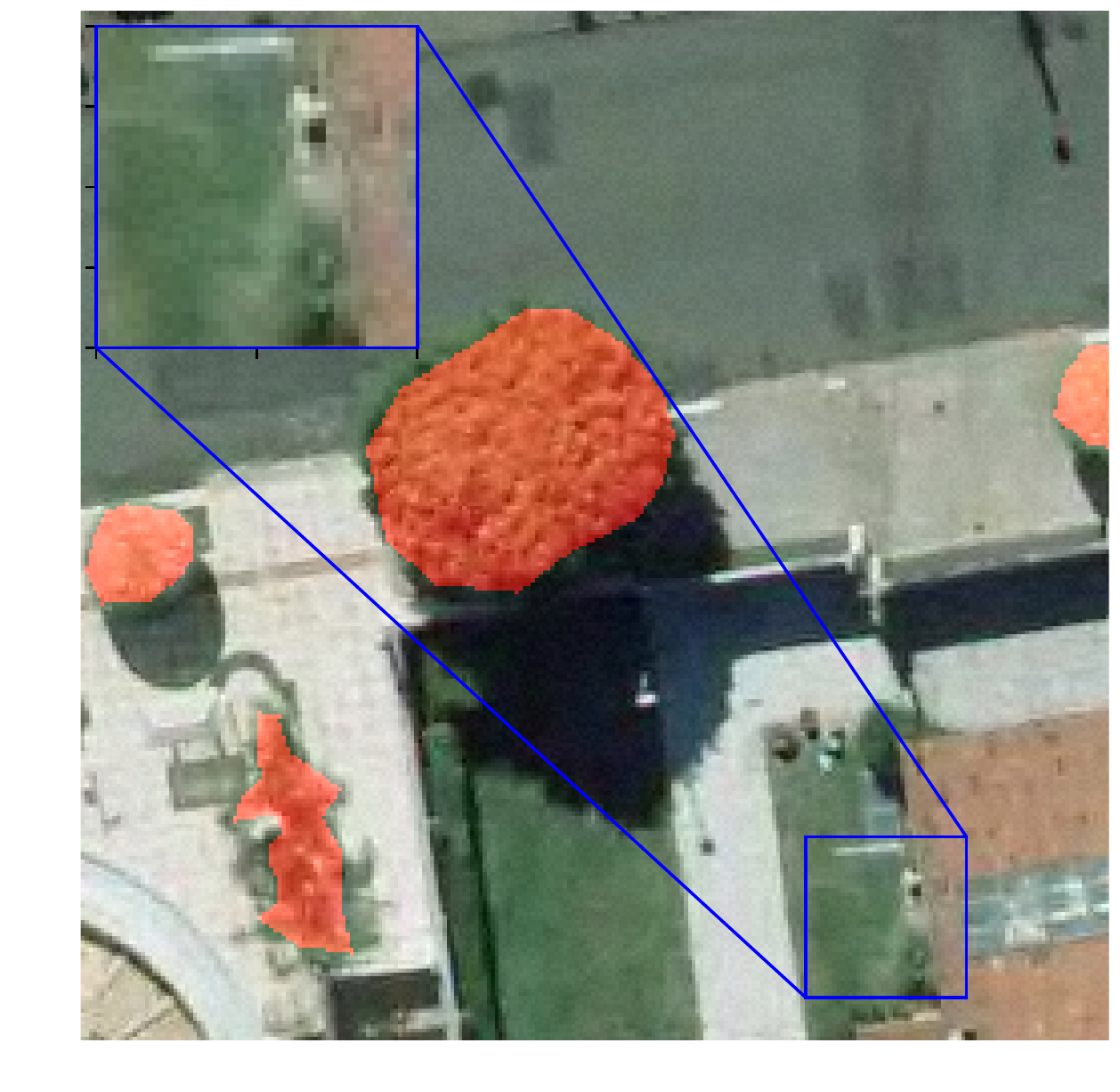}}
	\subfigure{\includegraphics[width=.32\columnwidth]{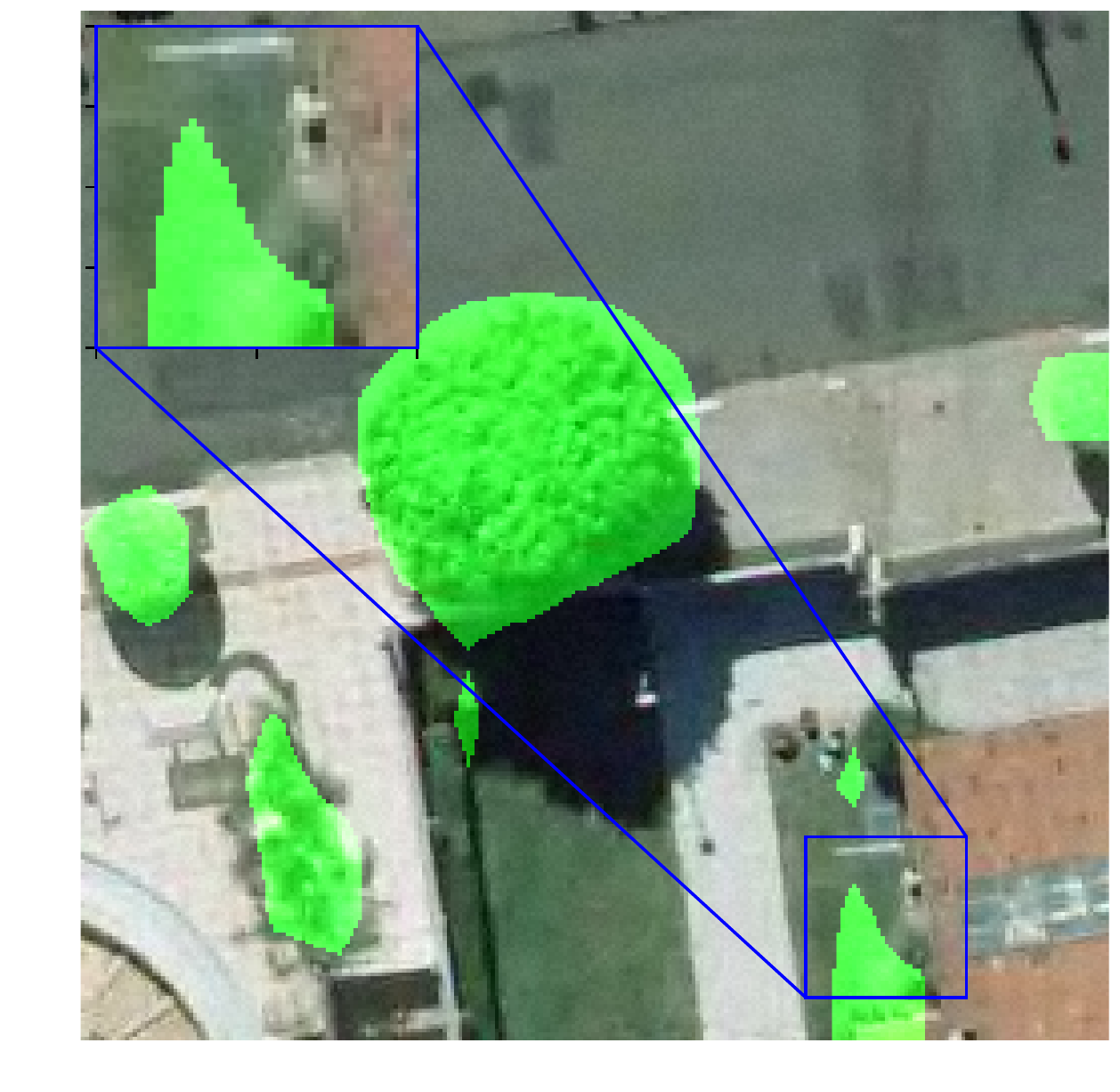}}
	\subfigure{\includegraphics[width=.32\columnwidth]{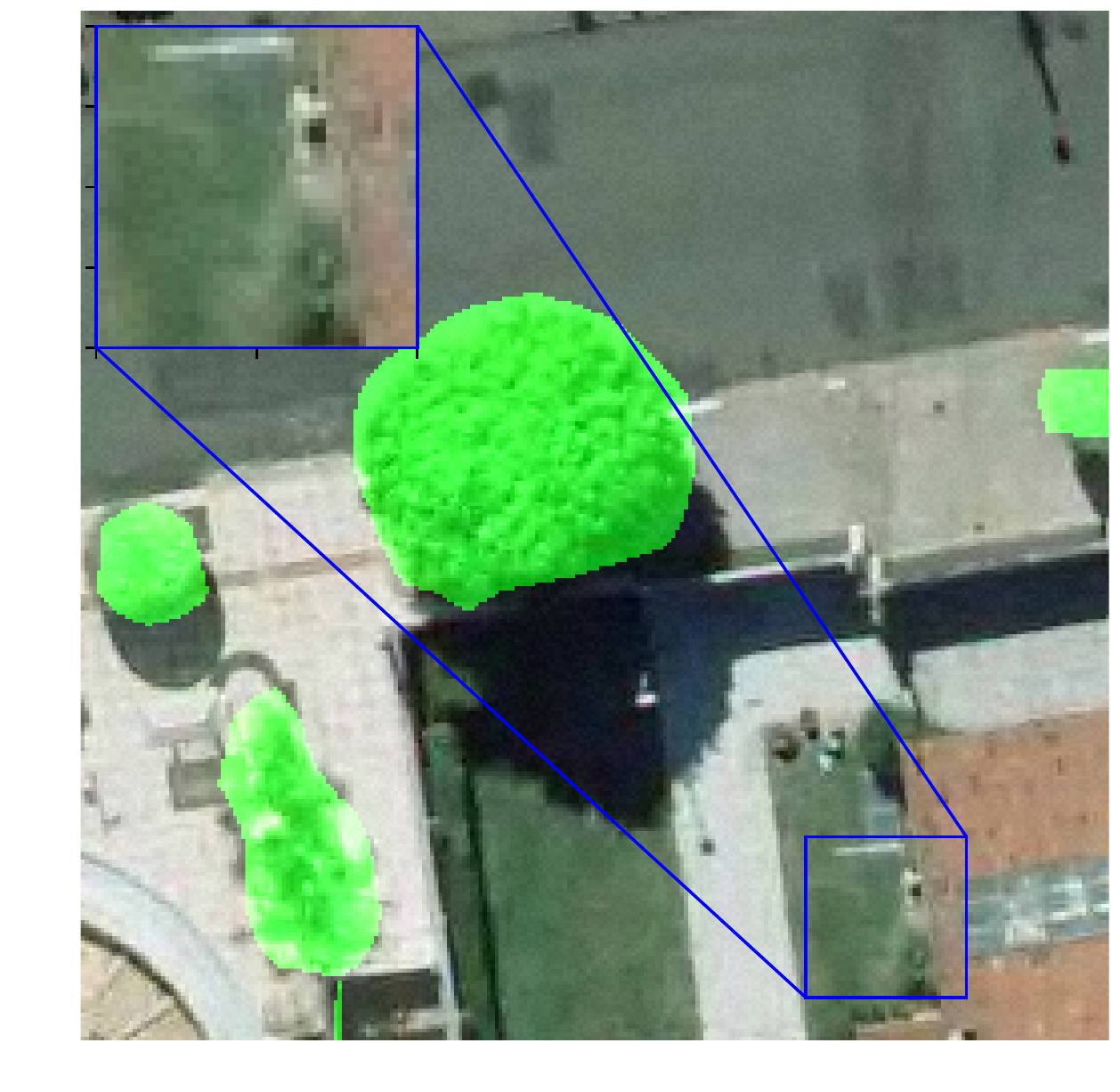}}

	\subfigure{\includegraphics[width=.32\columnwidth]{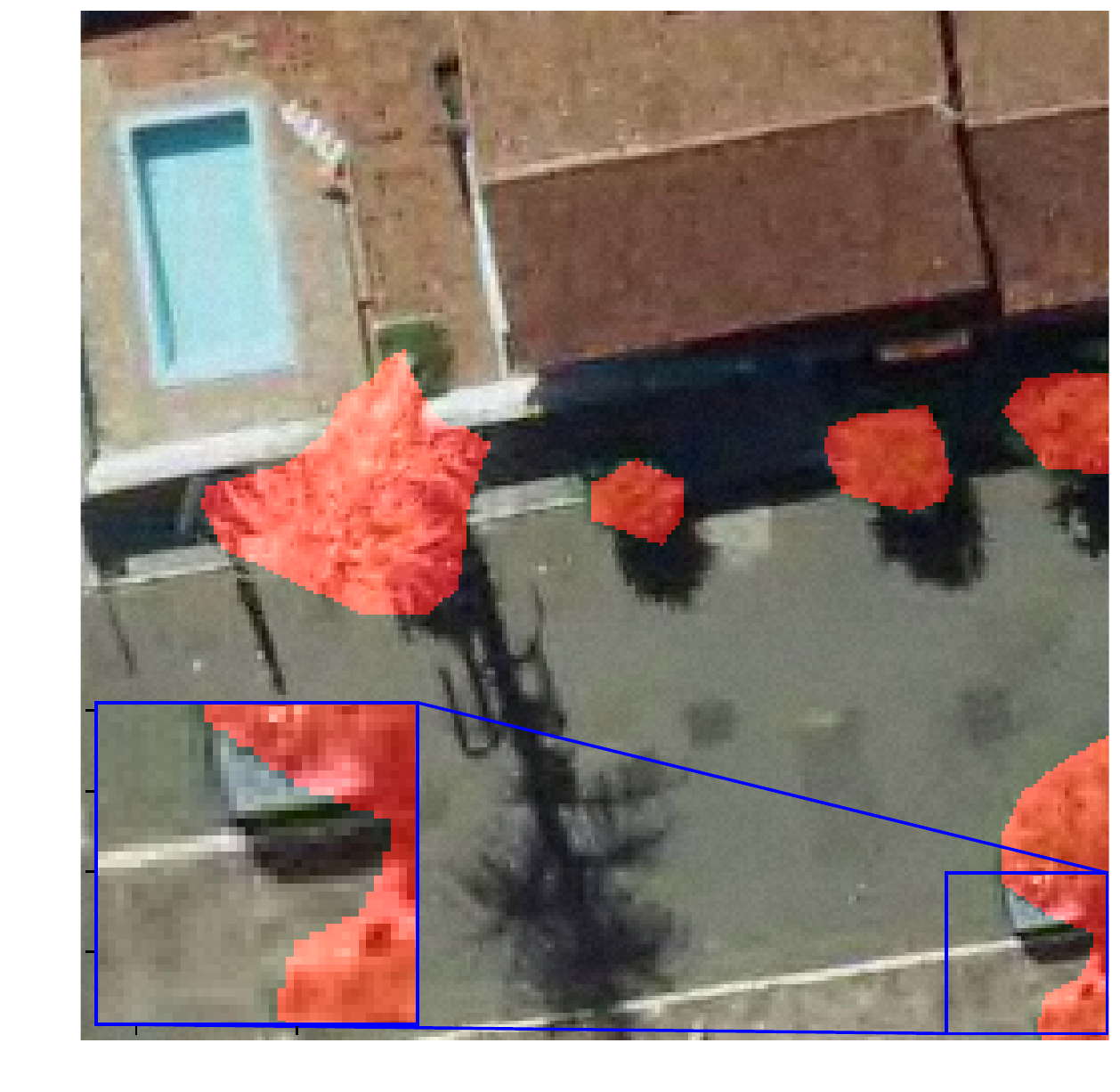}}
	\subfigure{\includegraphics[width=.32\columnwidth]{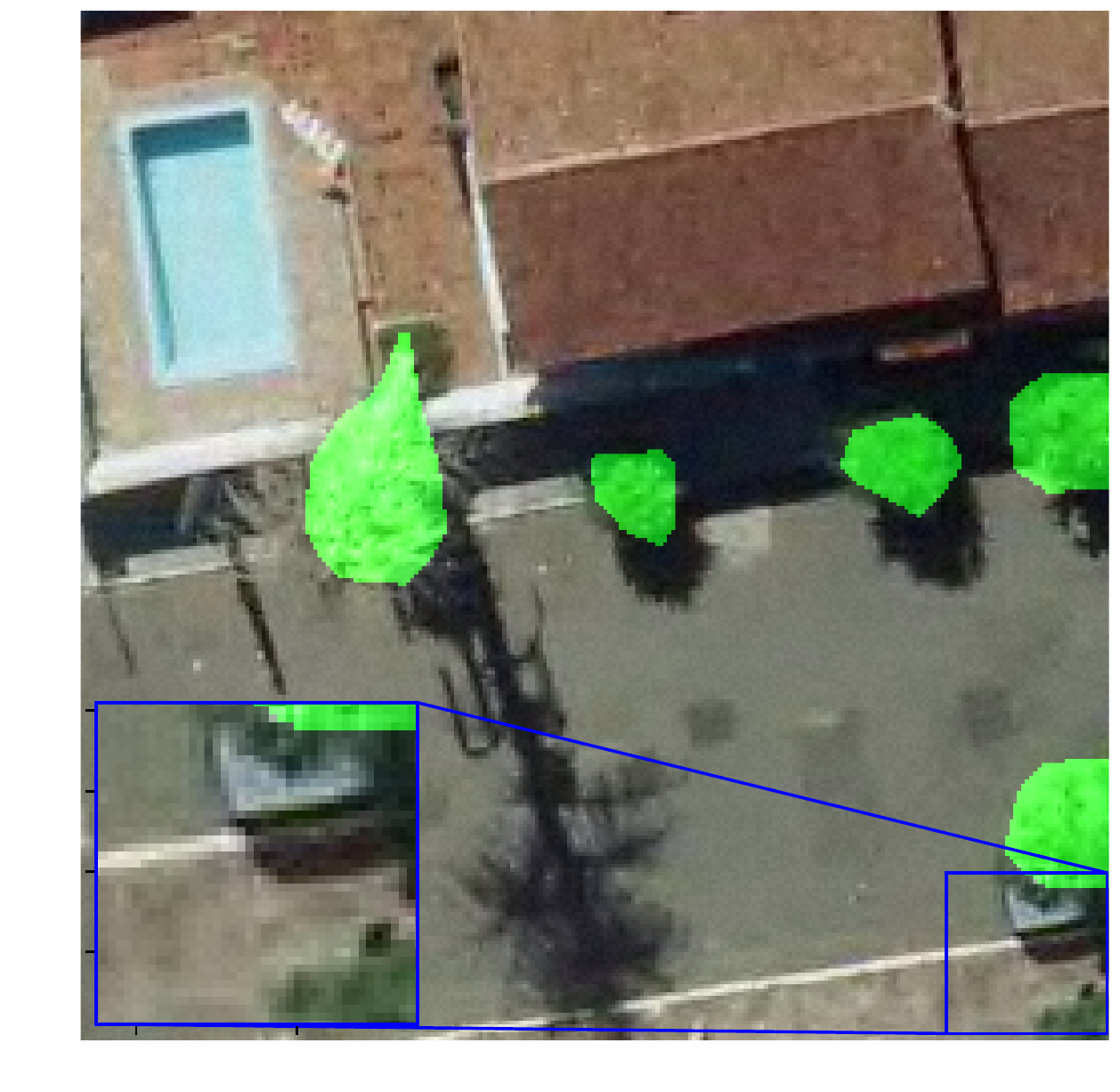}}
	\subfigure{\includegraphics[width=.32\columnwidth]{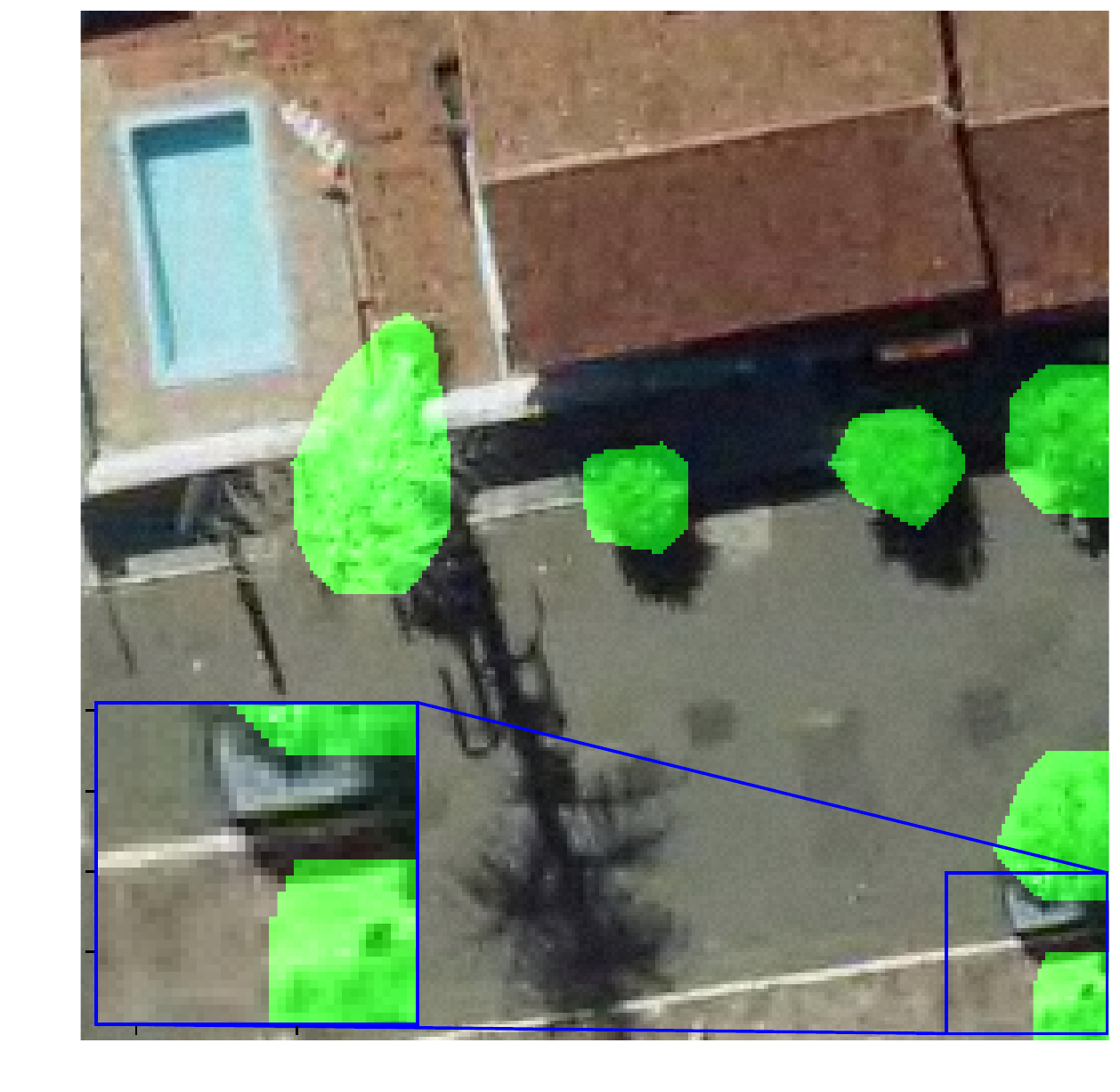}}
	\setcounter{subfigure}{0}
	\subfigure[Ground-truth]{\includegraphics[width=.32\columnwidth]{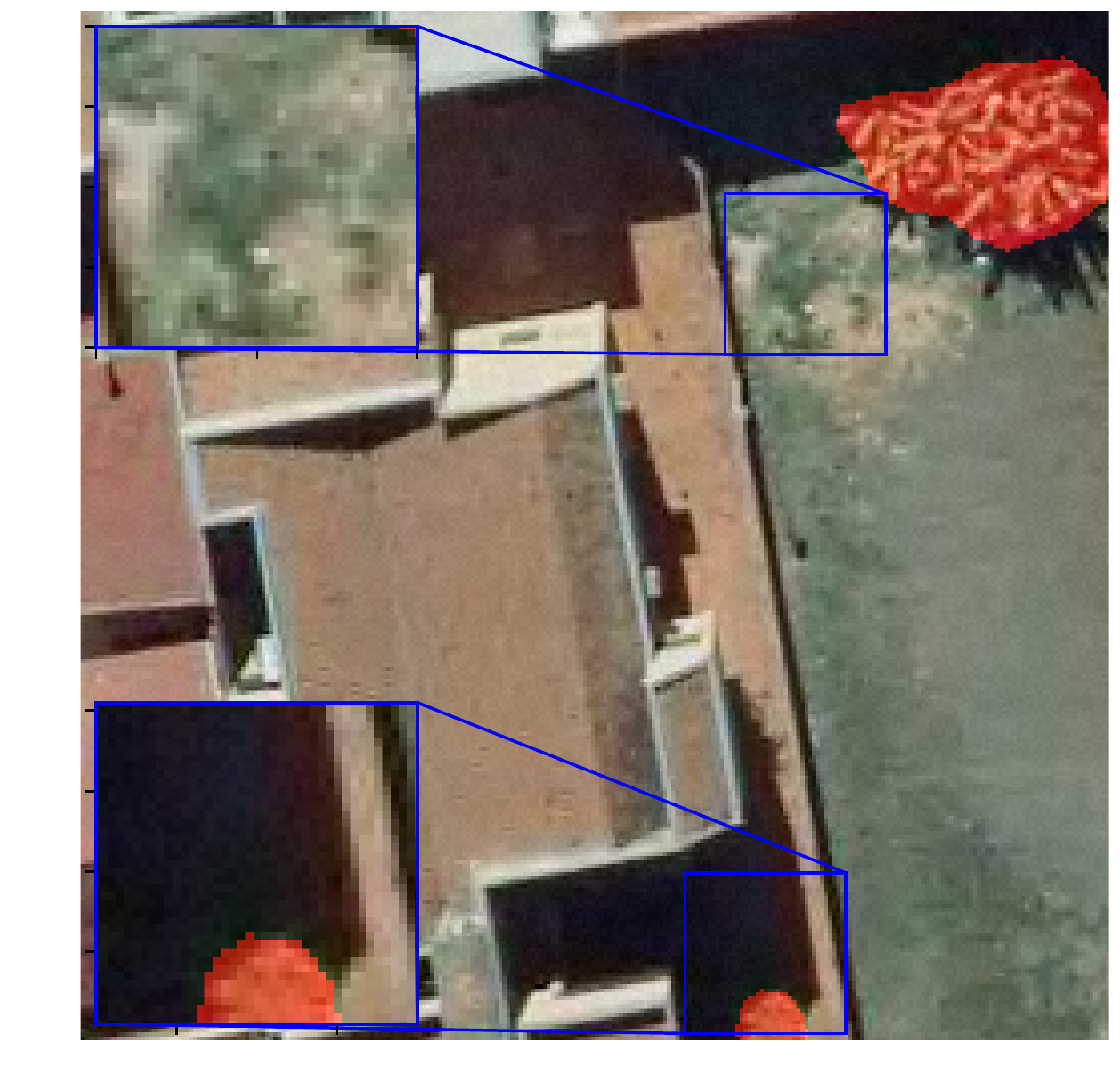}}
	\subfigure[FCN]{\includegraphics[width=.32\columnwidth]{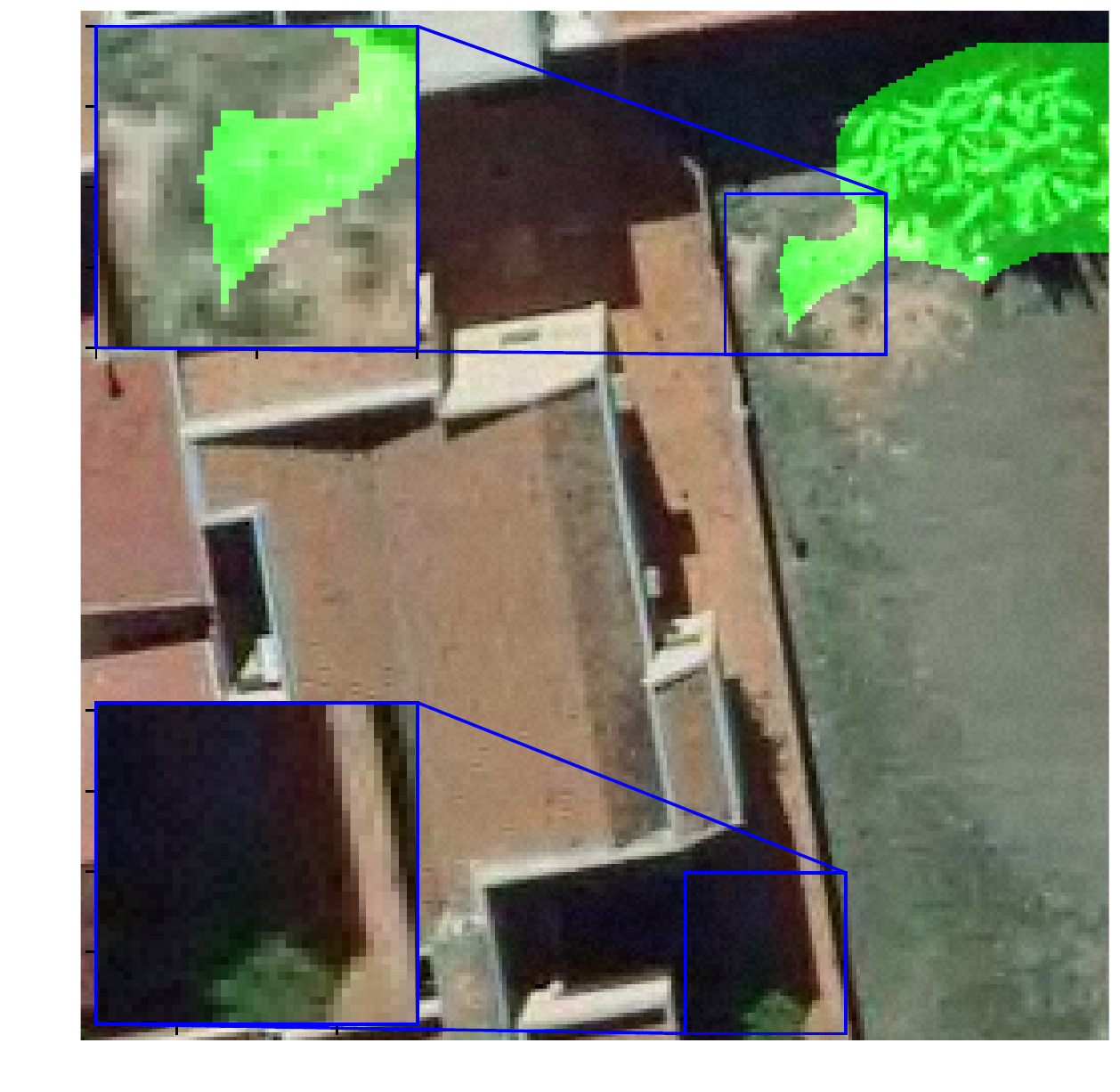}}
	\subfigure[Proposed Approach]{\includegraphics[width=.32\columnwidth]{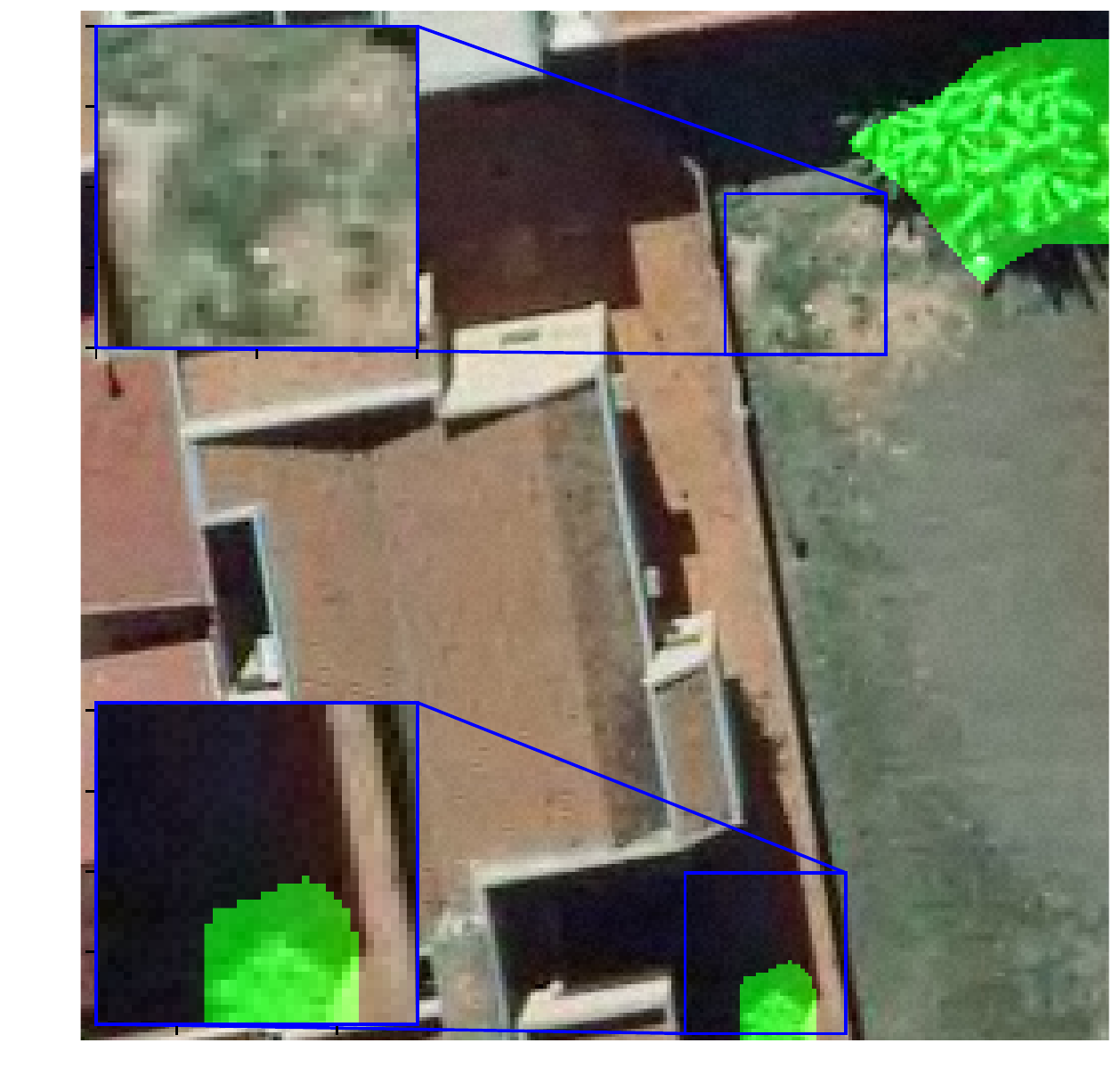}}
	
	\caption{\label{fig:tree_dataset} Example of (a) ground-truth, (b) FCN and (c) proposed approach from Urban Tree dataset.}
\end{figure}

\textbf{Urban tree dataset.}
Fig. \ref{fig:tree_dataset} presents three examples that show the advantage of the proposed approach. The first column shows the ground-truth while the second and third columns present the result of the segmentation using the baseline and the proposed approach. The first example (first row) shows that the baseline incorrectly segments the grass as a tree. On the other hand, the proposed approach can correctly segment the grass as a background, even though the colors are similar. The second example shows that the proposed approach is capable of correctly segmenting small foreground regions. This is because the importance of these pixels is increased during training and the weights of the convolutional layers tend to adjust better for these regions. Finally, the third example also shows small regions correctly segmented by the proposed approach. Also, it is possible to observe that the tree edge is better defined when compared to the baseline. This is possible due to the uncertainty included in tree-border regions, which are hardly labeled correctly. Concerning the border of objects, the proposed method decreases the importance of pixels, making CNN weights take this into account.

\begin{figure}[!ht]
	\centering

	\subfigure{\includegraphics[width=.32\columnwidth]{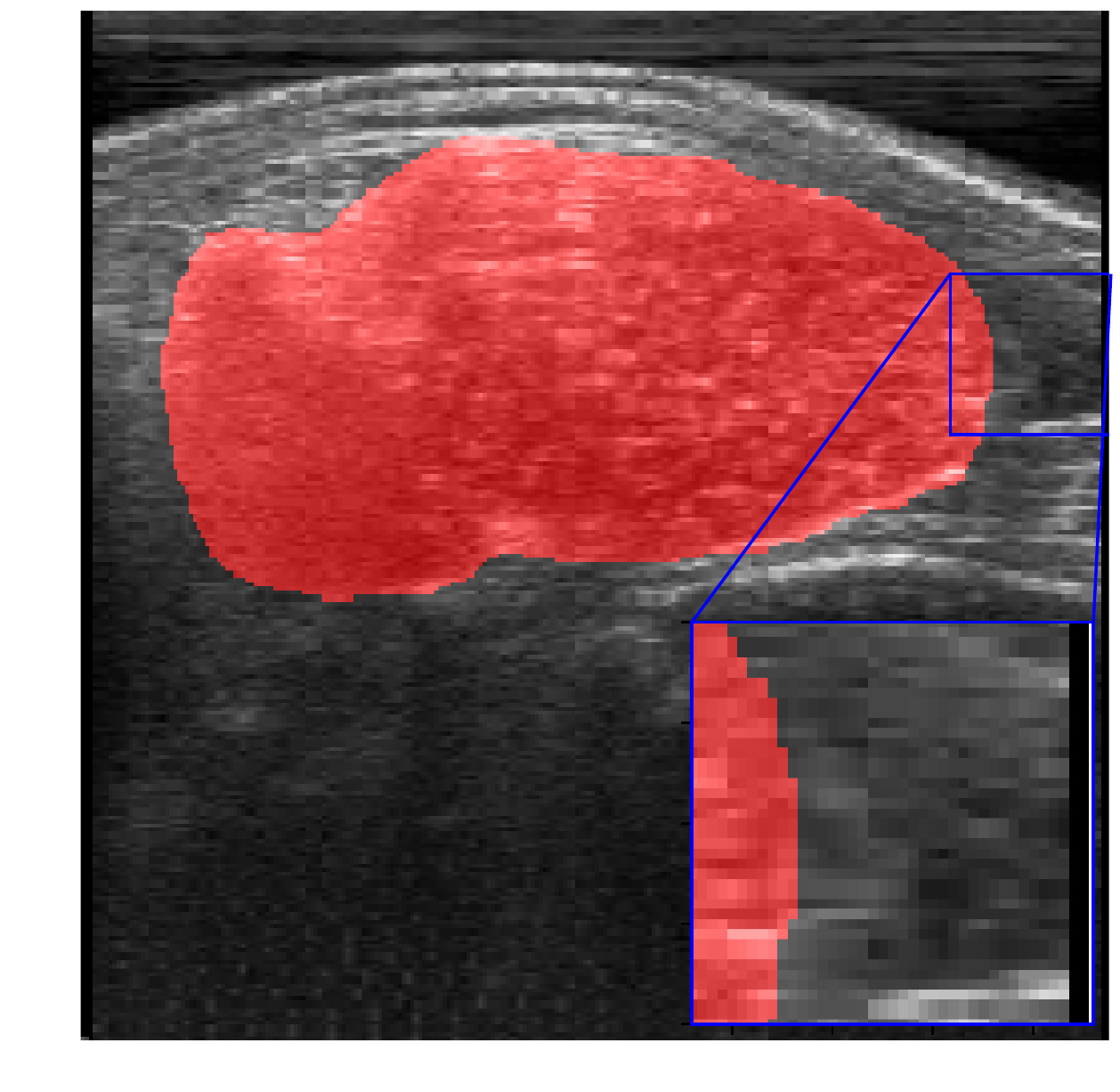}}
	\subfigure{\includegraphics[width=.32\columnwidth]{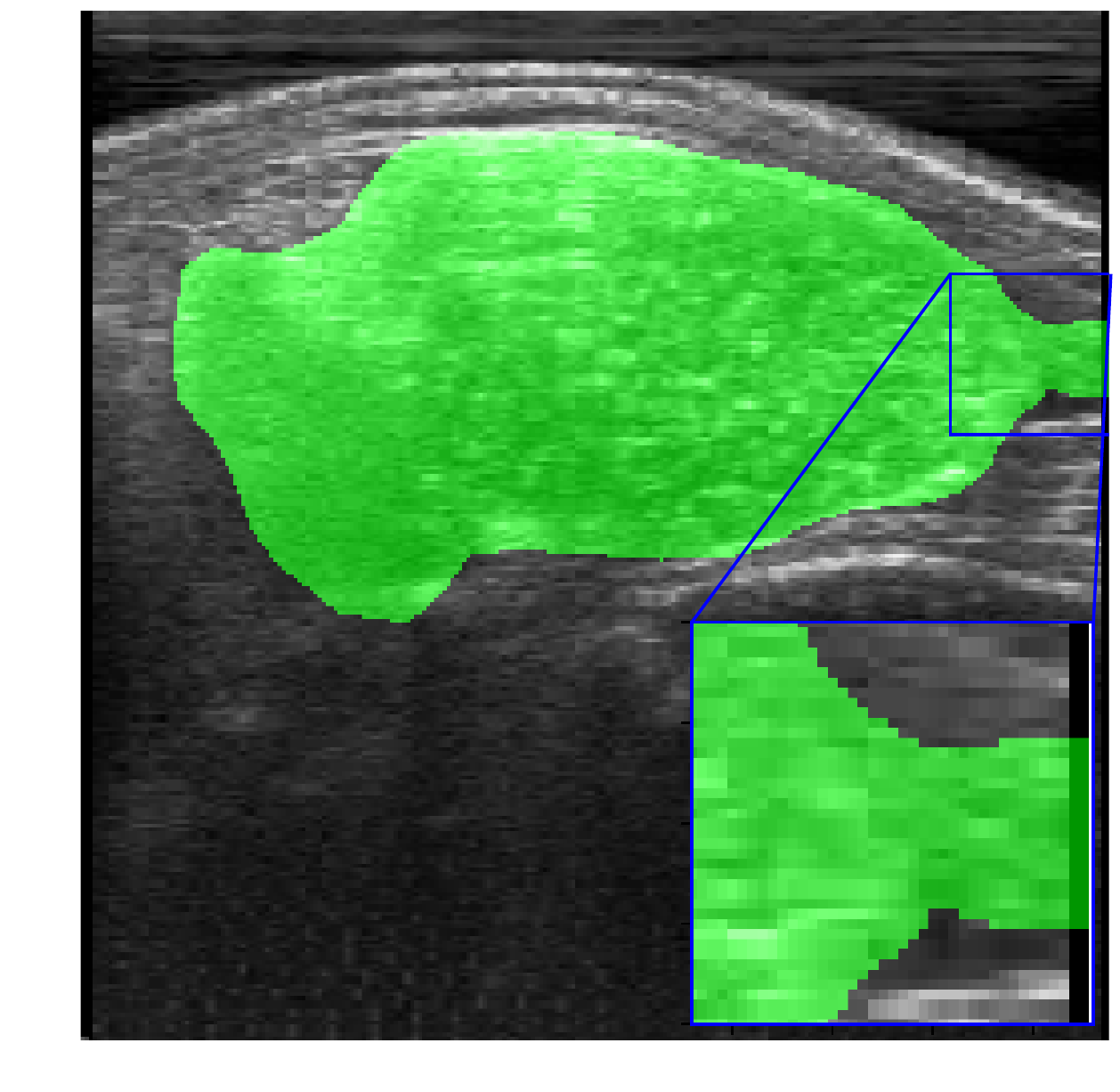}}
	\subfigure{\includegraphics[width=.32\columnwidth]{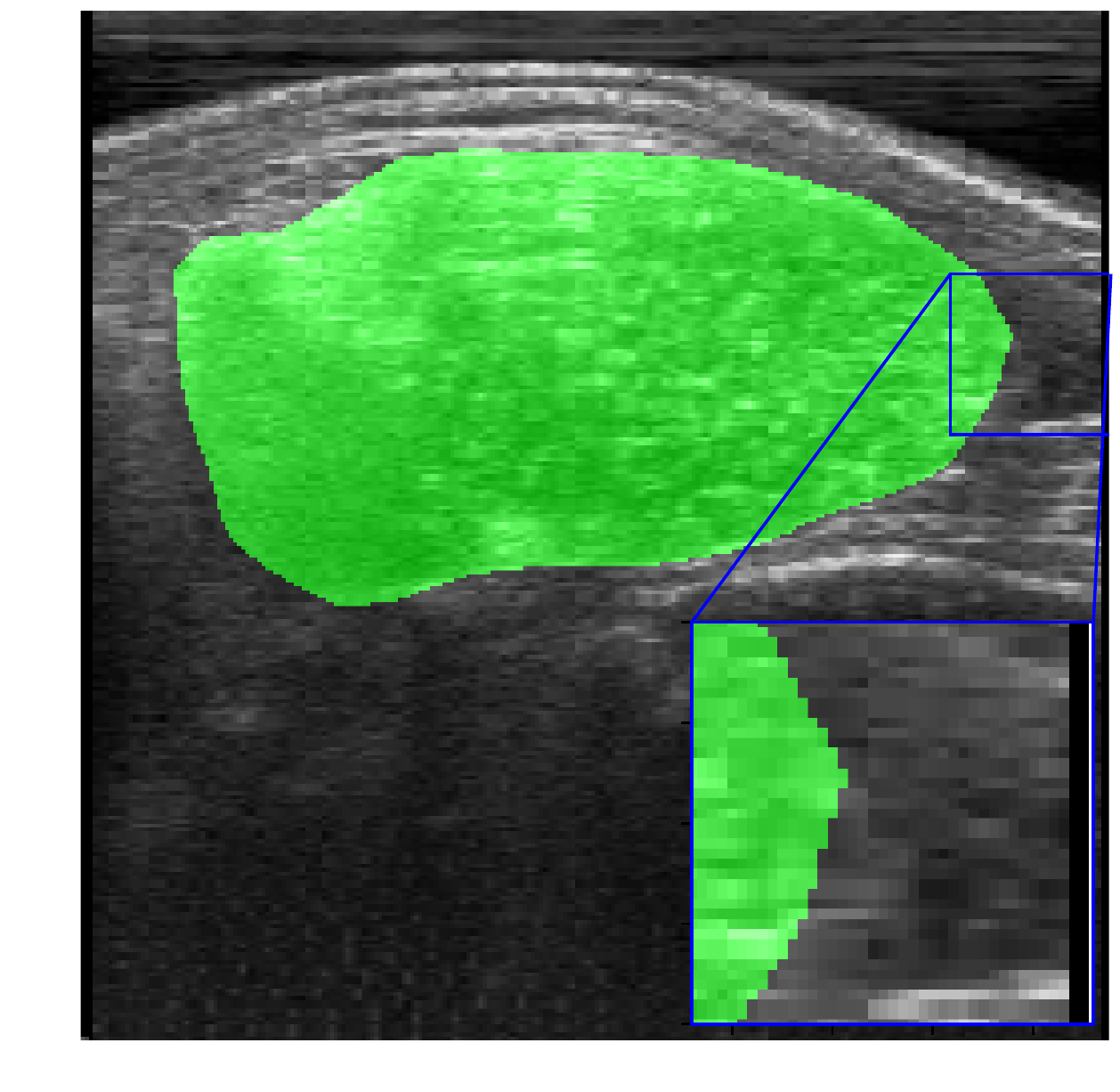}}
	\subfigure{\includegraphics[width=.32\columnwidth]{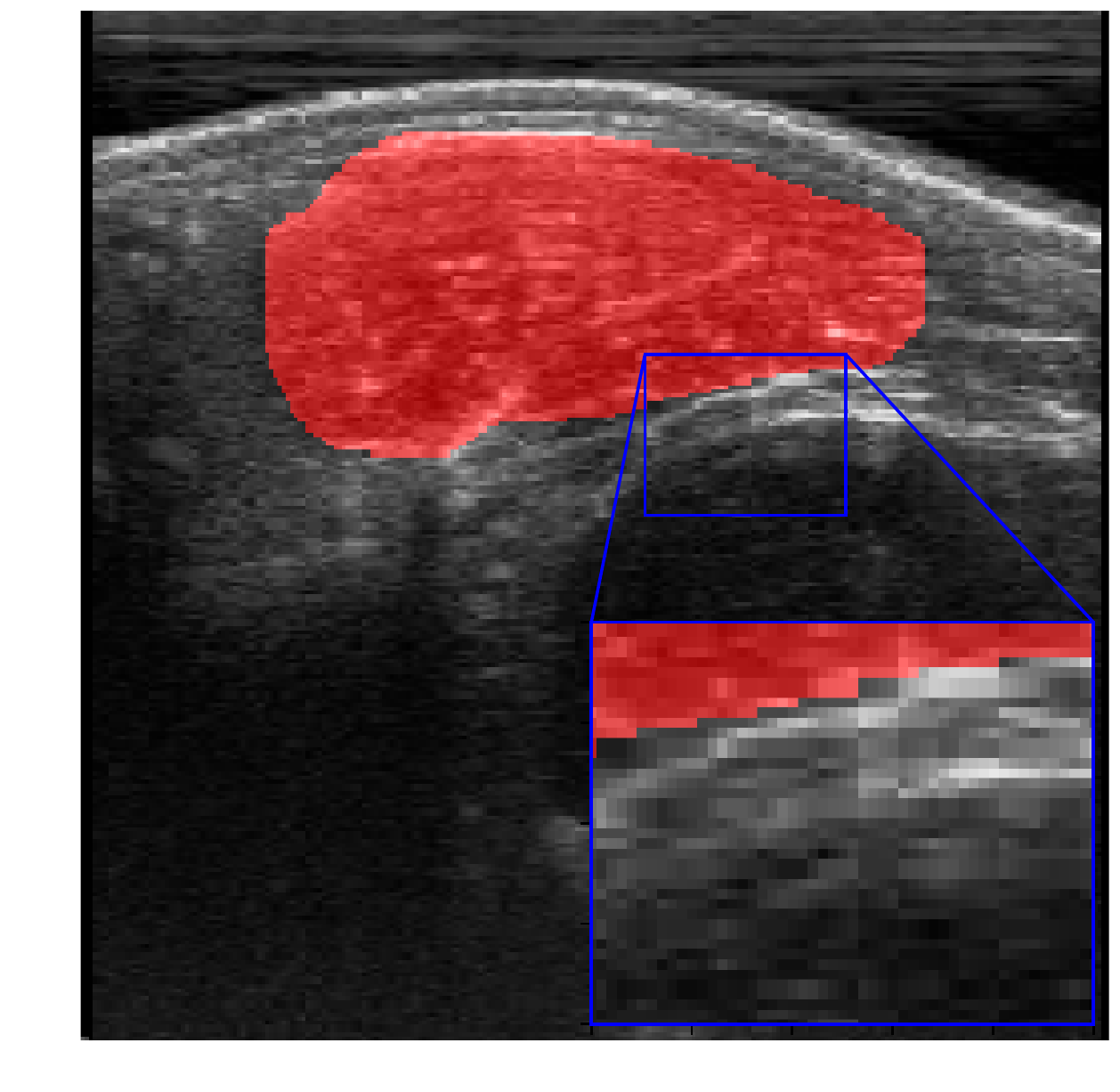}}
	\subfigure{\includegraphics[width=.32\columnwidth]{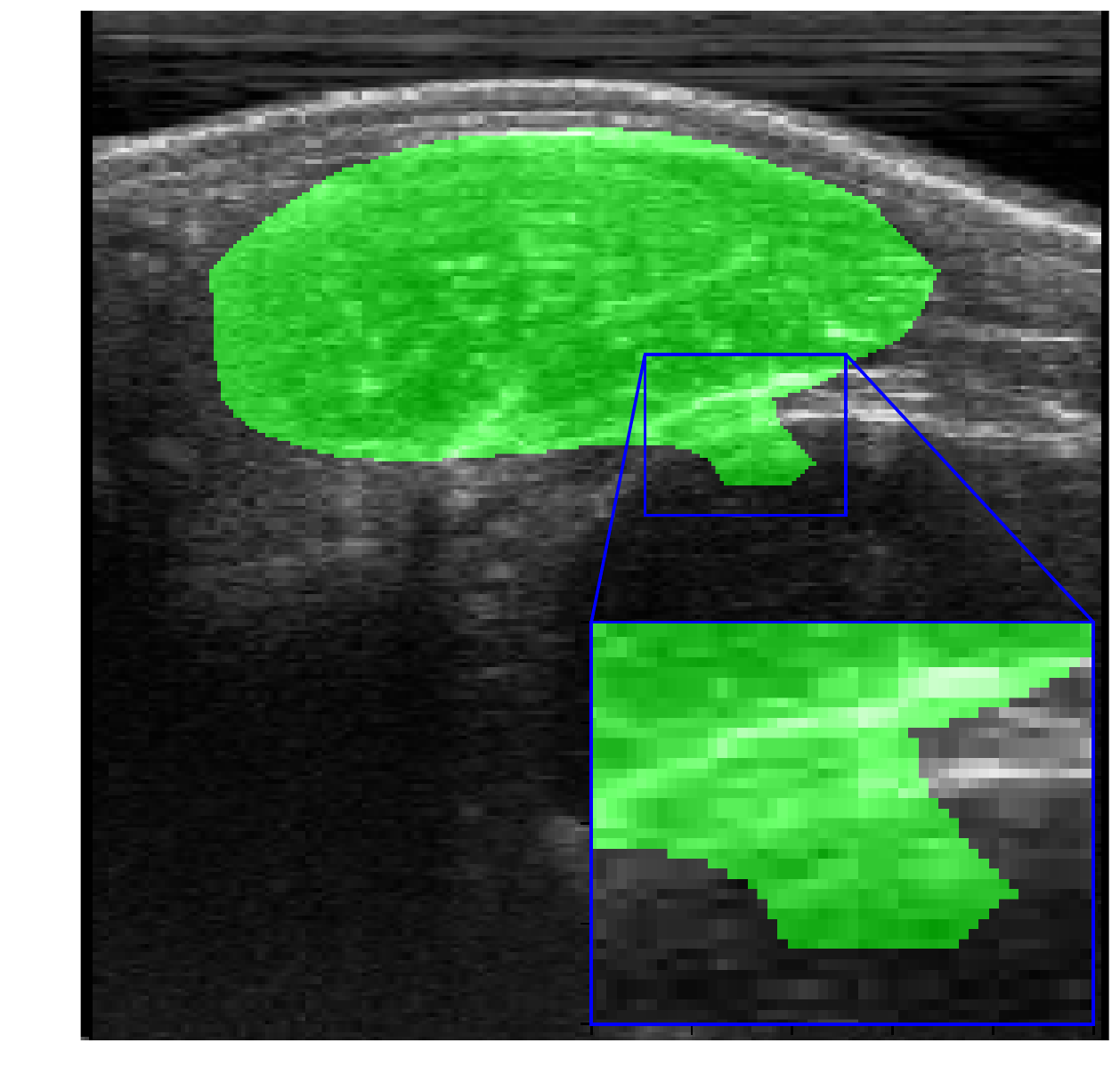}}
	\subfigure{\includegraphics[width=.32\columnwidth]{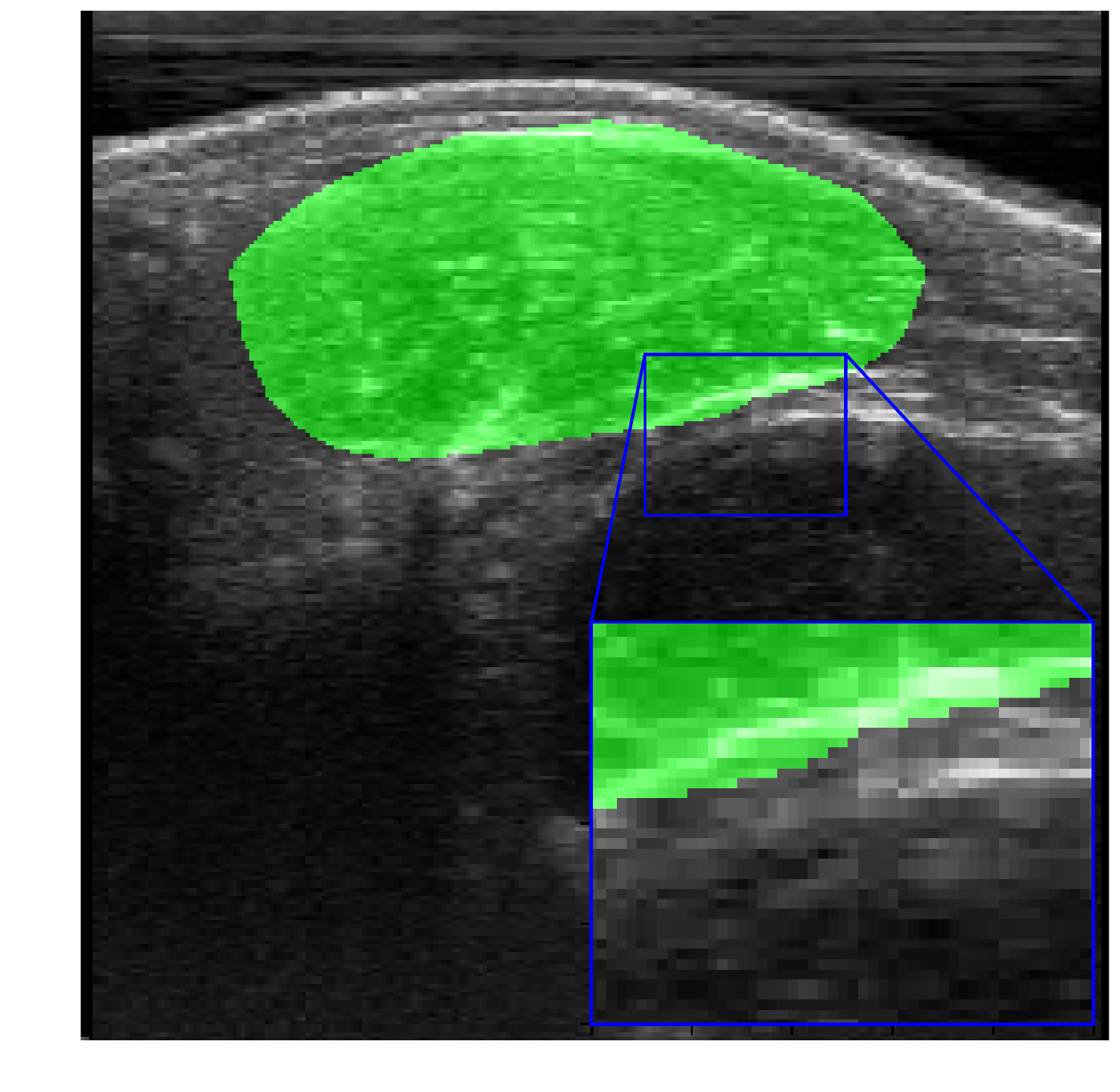}}
	
	\setcounter{subfigure}{0}
	\subfigure[Ground-truth]{\includegraphics[width=.32\columnwidth]{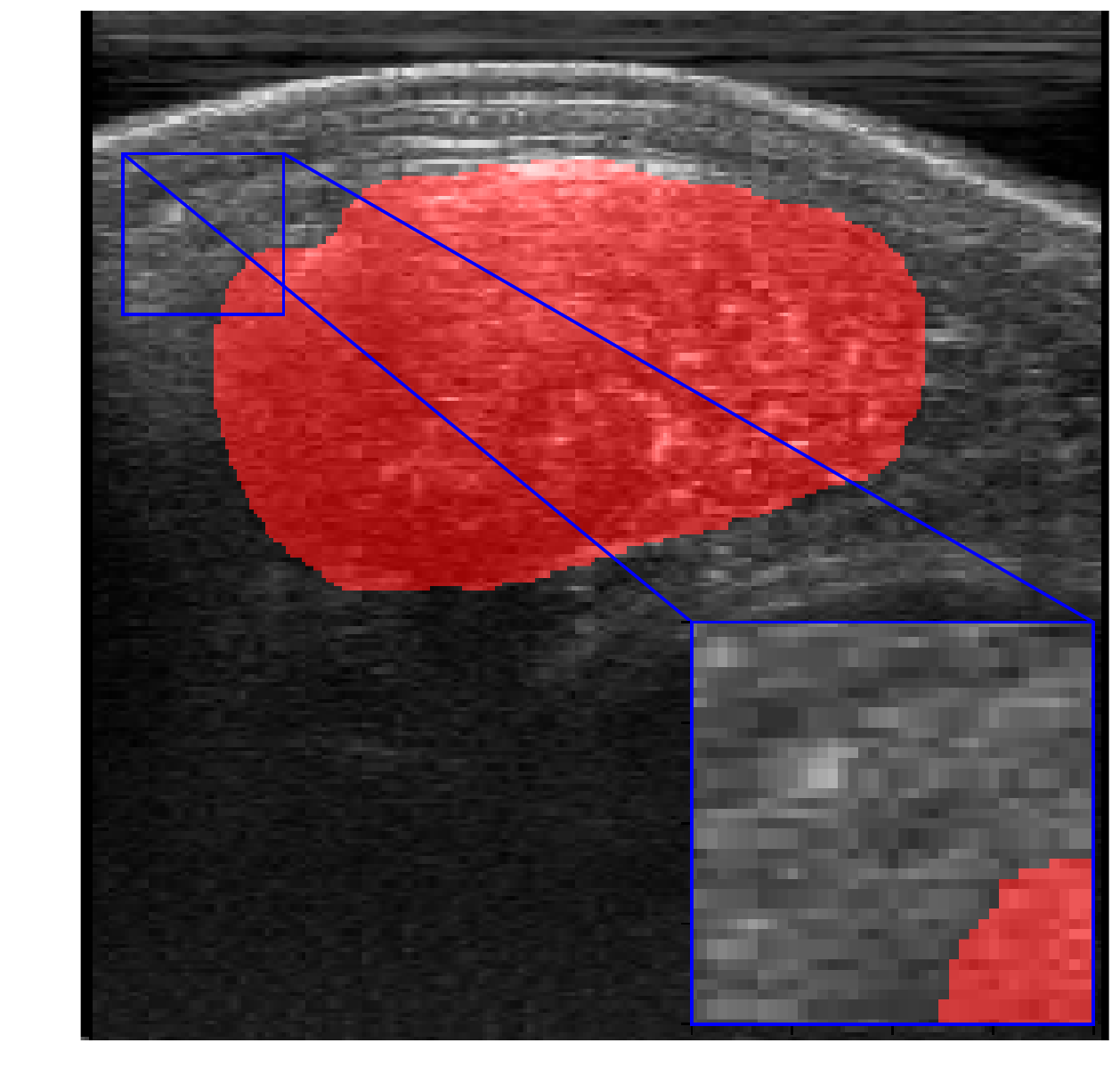}}
	\subfigure[FCN]{\includegraphics[width=.32\columnwidth]{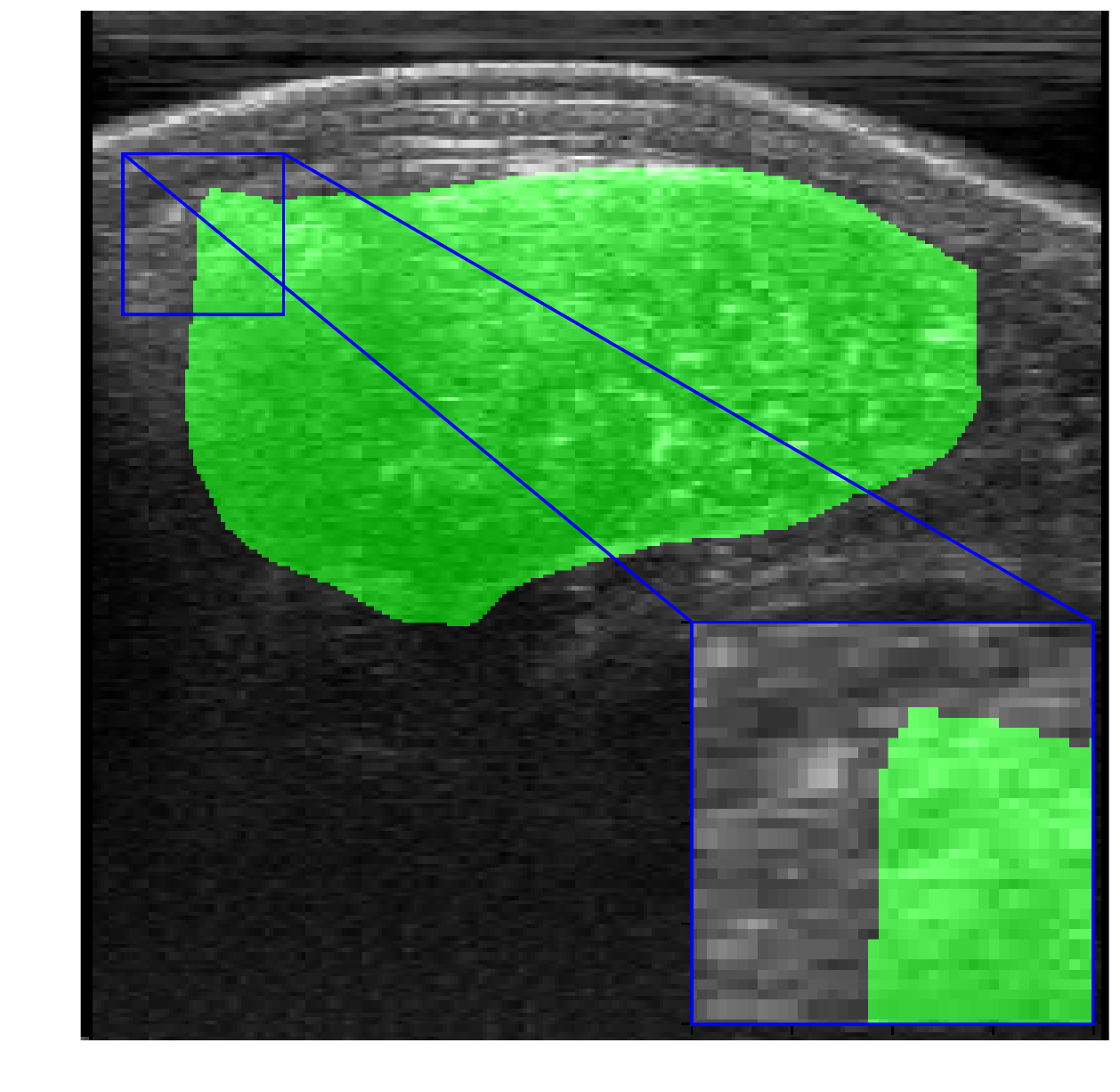}}
	\subfigure[Proposed Approach]{\includegraphics[width=.32\columnwidth]{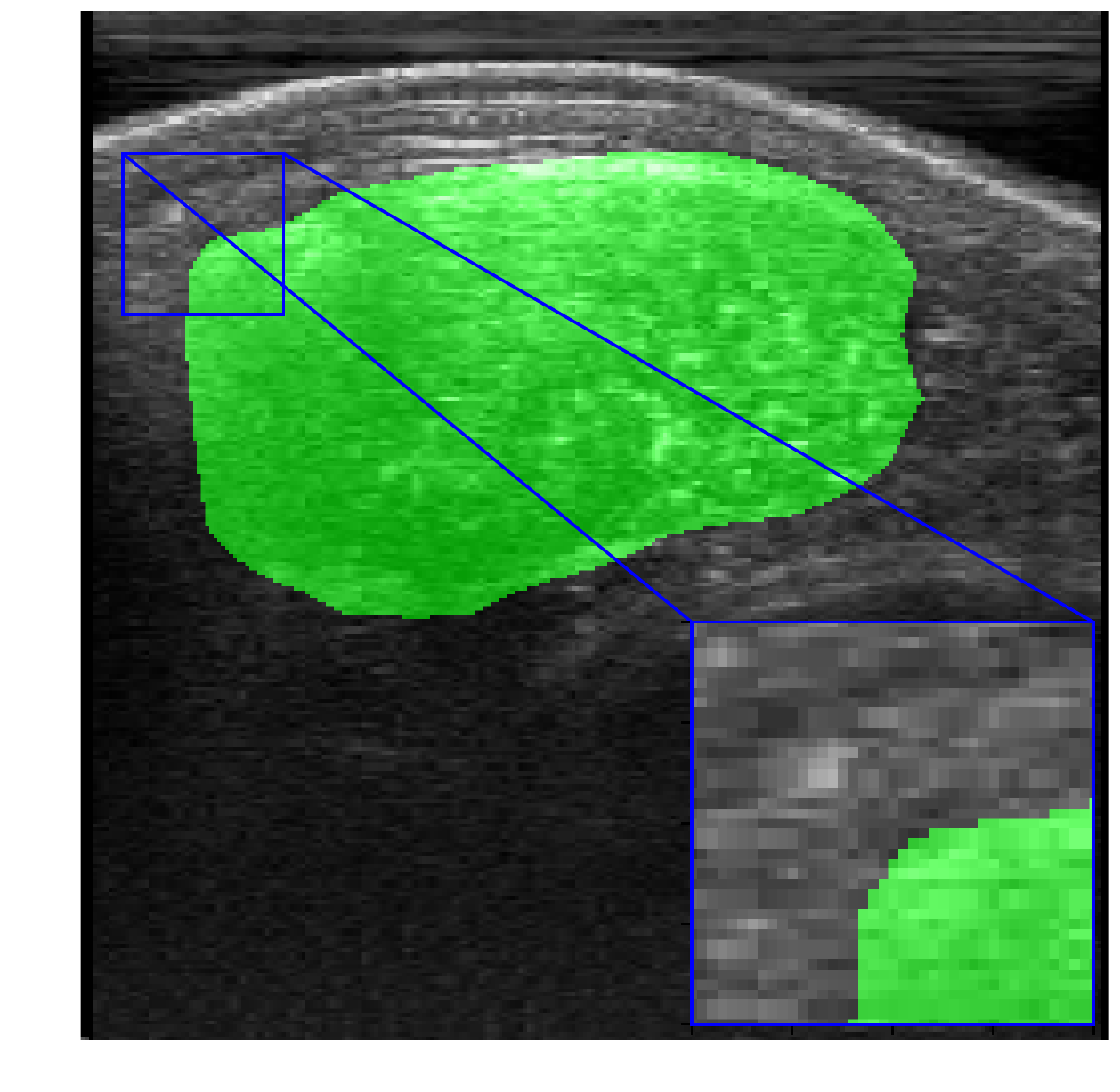}}
	
	\caption{\label{fig:REA_dataset} Example of (a) ground-truth, (b) FCN and (c) proposed approach from REA dataset.}
\end{figure}

\textbf{REA dataset.}
This image dataset has high uncertainty during labeling due to noise from the ultrasound image.
In some cases, the border of REA is not completely visible and must be estimated by the specialist.
Therefore, the proposed approach becomes essential to obtain accurate segmentation at the edges.
The segmentation examples in Fig. \ref{fig:REA_dataset} show that the baseline was not able to define the REA correctly due to the uncertainty of the labeling. On the other hand, the proposed approach presents results close to the specialist in regions that the border needs to be estimated.

\begin{figure}[!ht]
	\centering
	\subfigure{\includegraphics[width=.32\columnwidth]{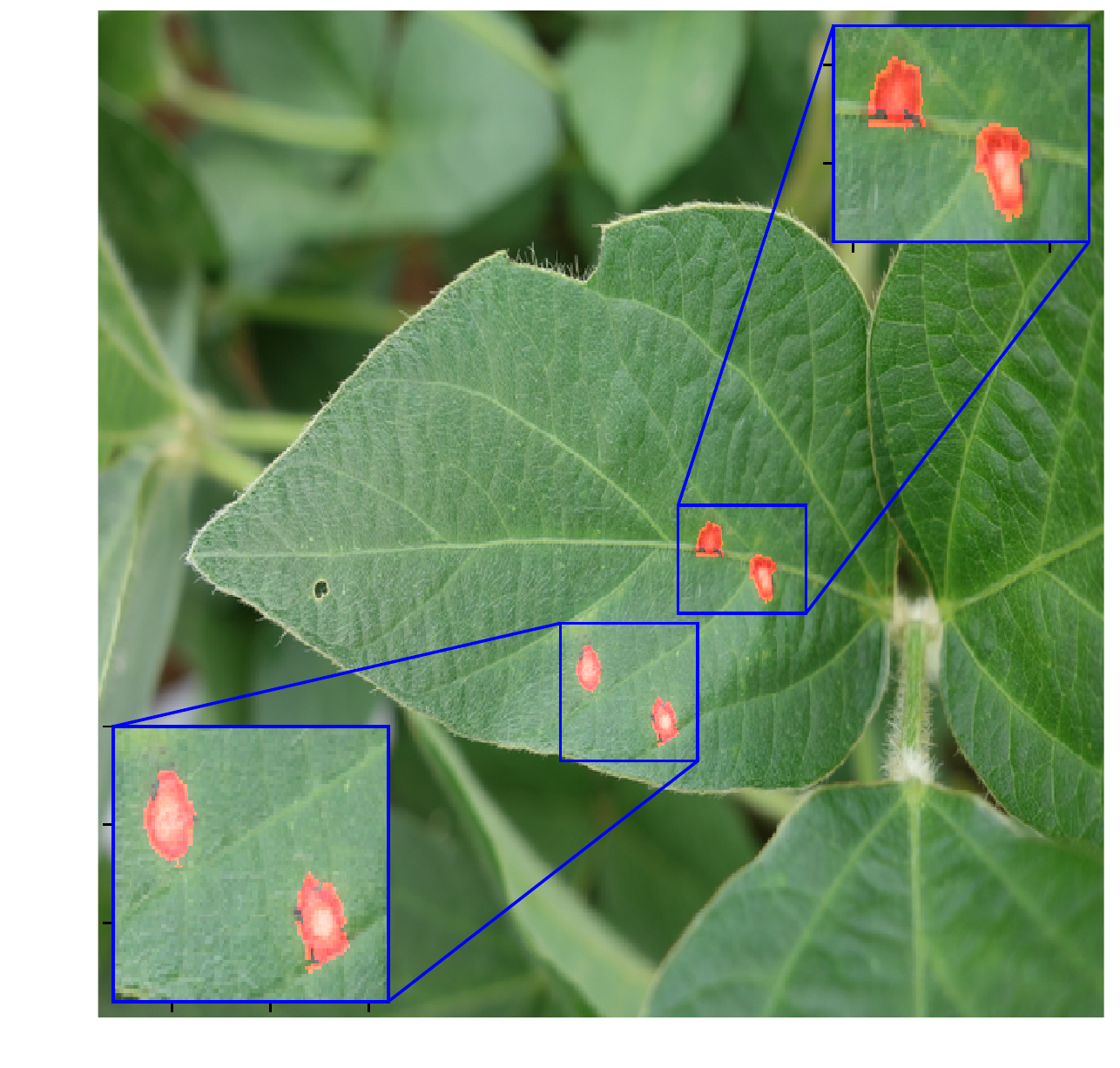}}
	\subfigure{\includegraphics[width=.32\columnwidth]{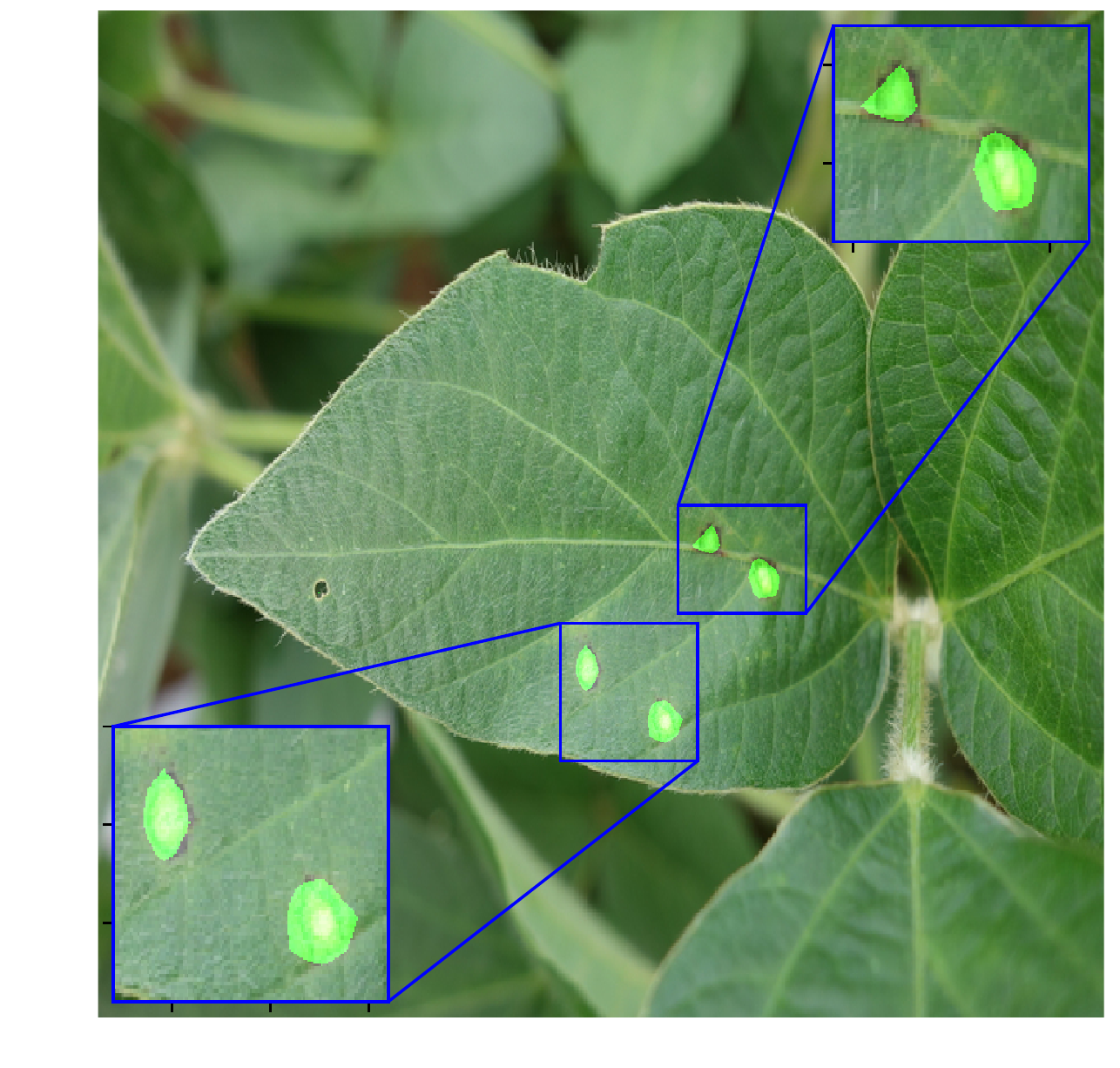}}
	\subfigure{\includegraphics[width=.32\columnwidth]{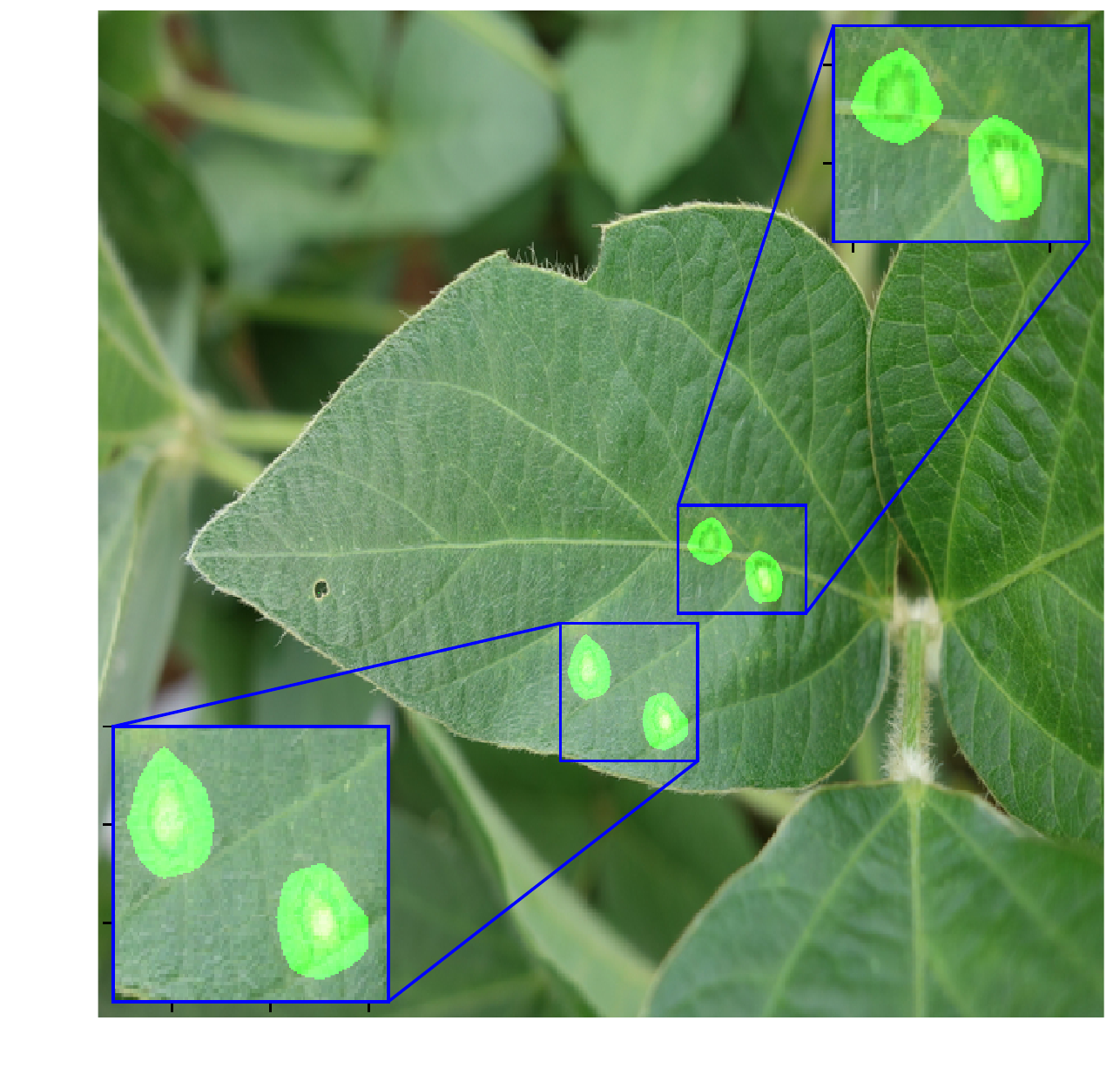}}
	
	\subfigure{\includegraphics[width=.32\columnwidth]{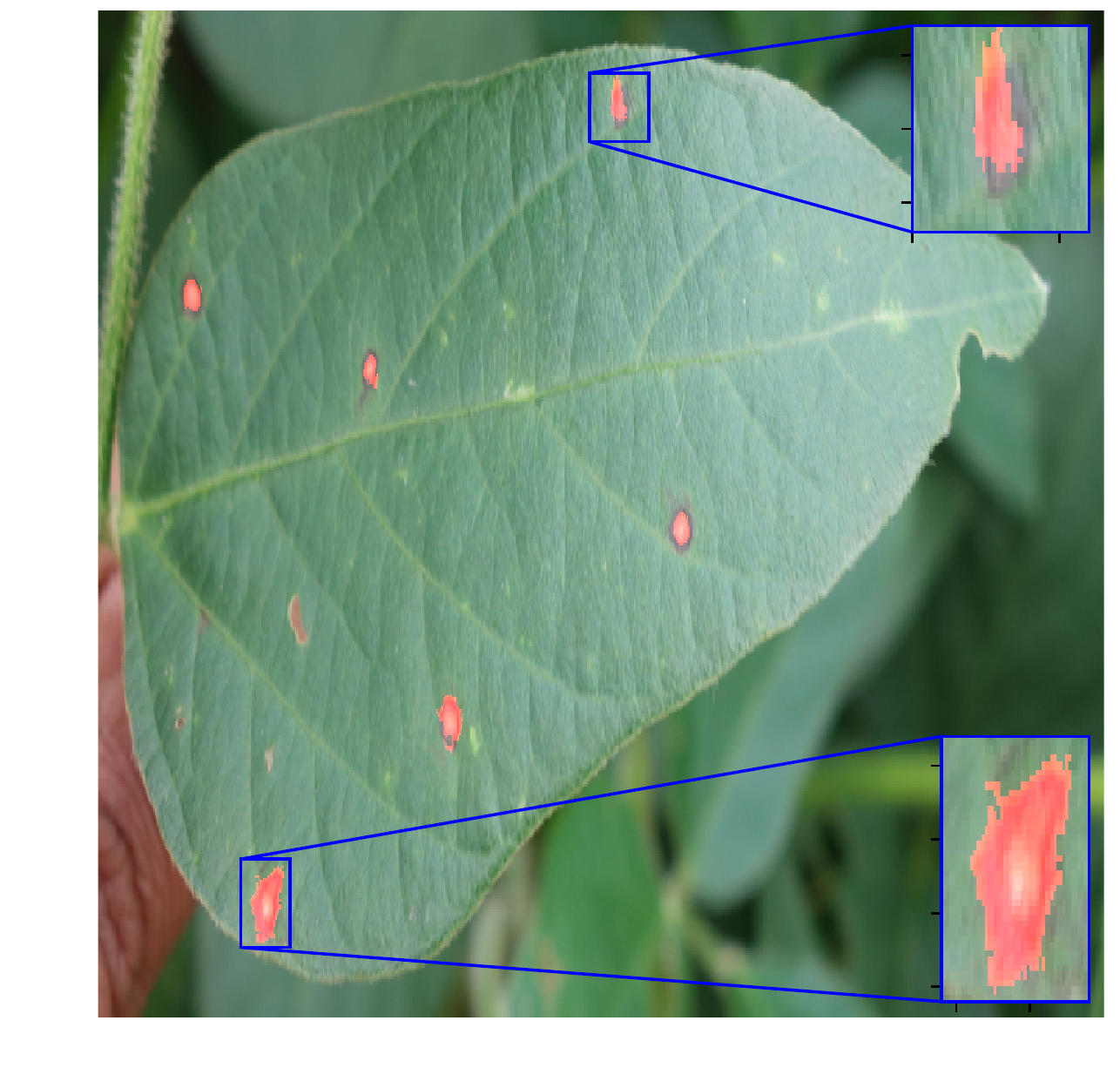}}
	\subfigure{\includegraphics[width=.32\columnwidth]{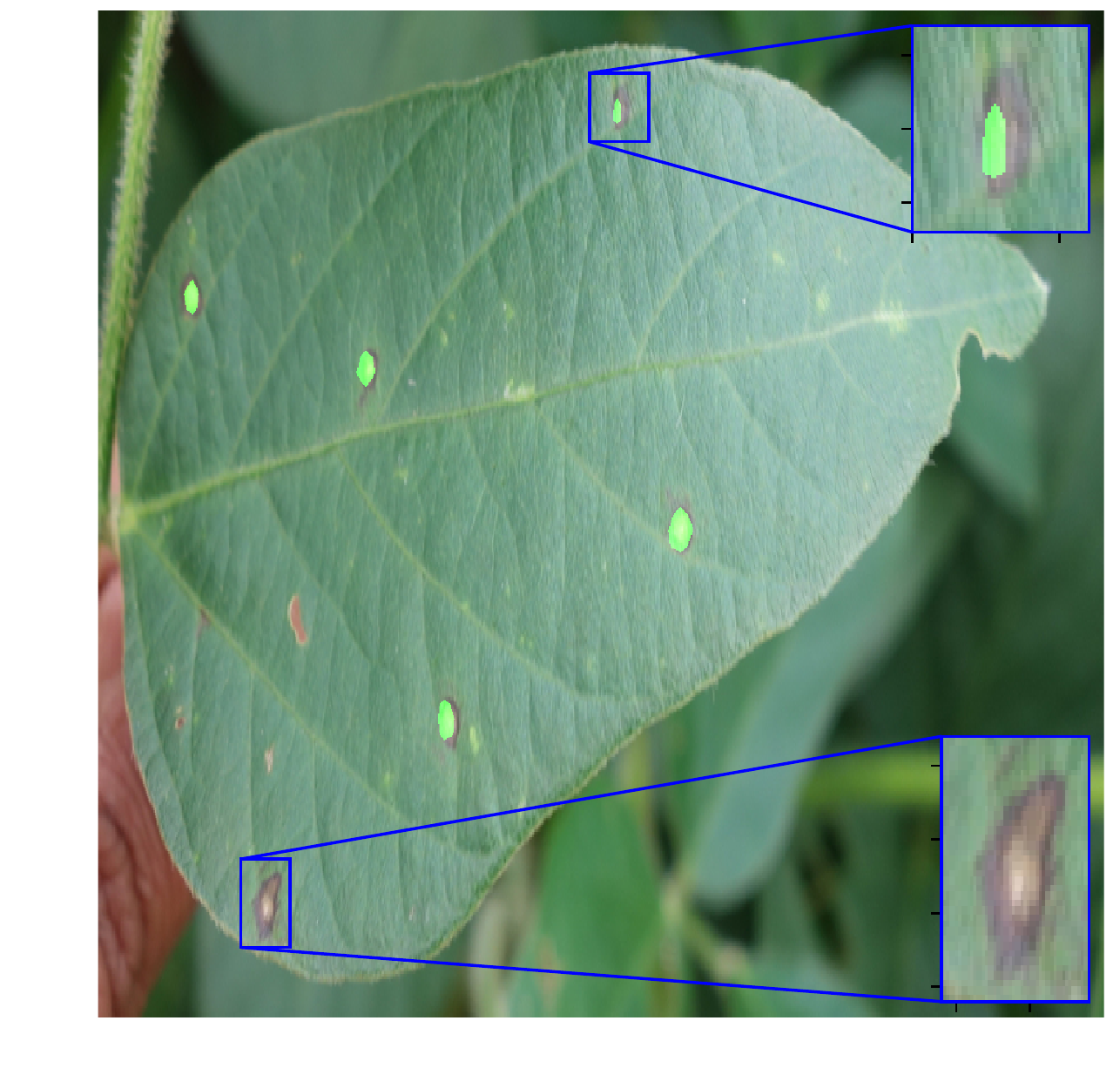}}
	\subfigure{\includegraphics[width=.32\columnwidth]{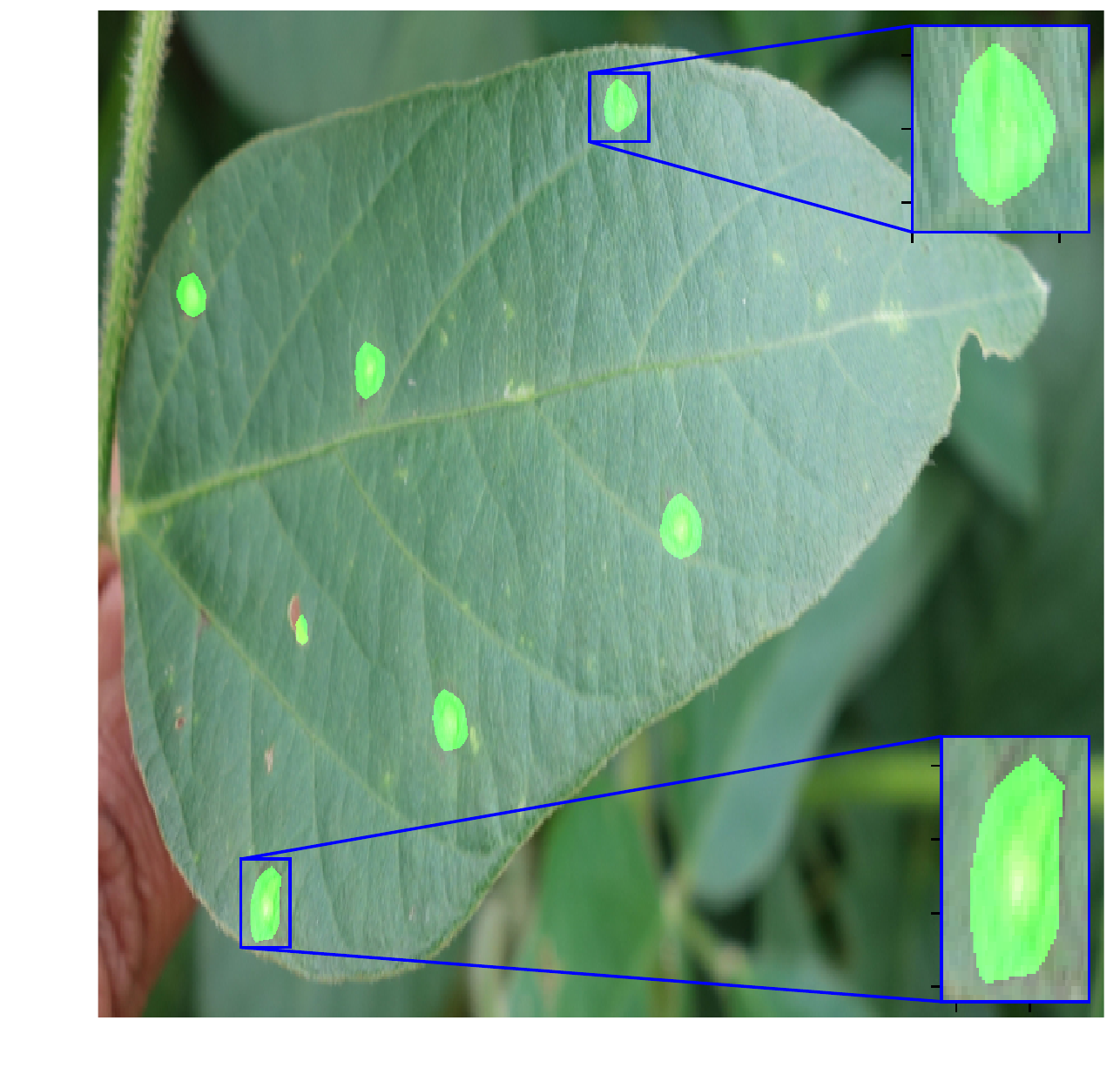}}
	
	\setcounter{subfigure}{0}
	\subfigure[Ground-truth]{\includegraphics[width=.32\columnwidth]{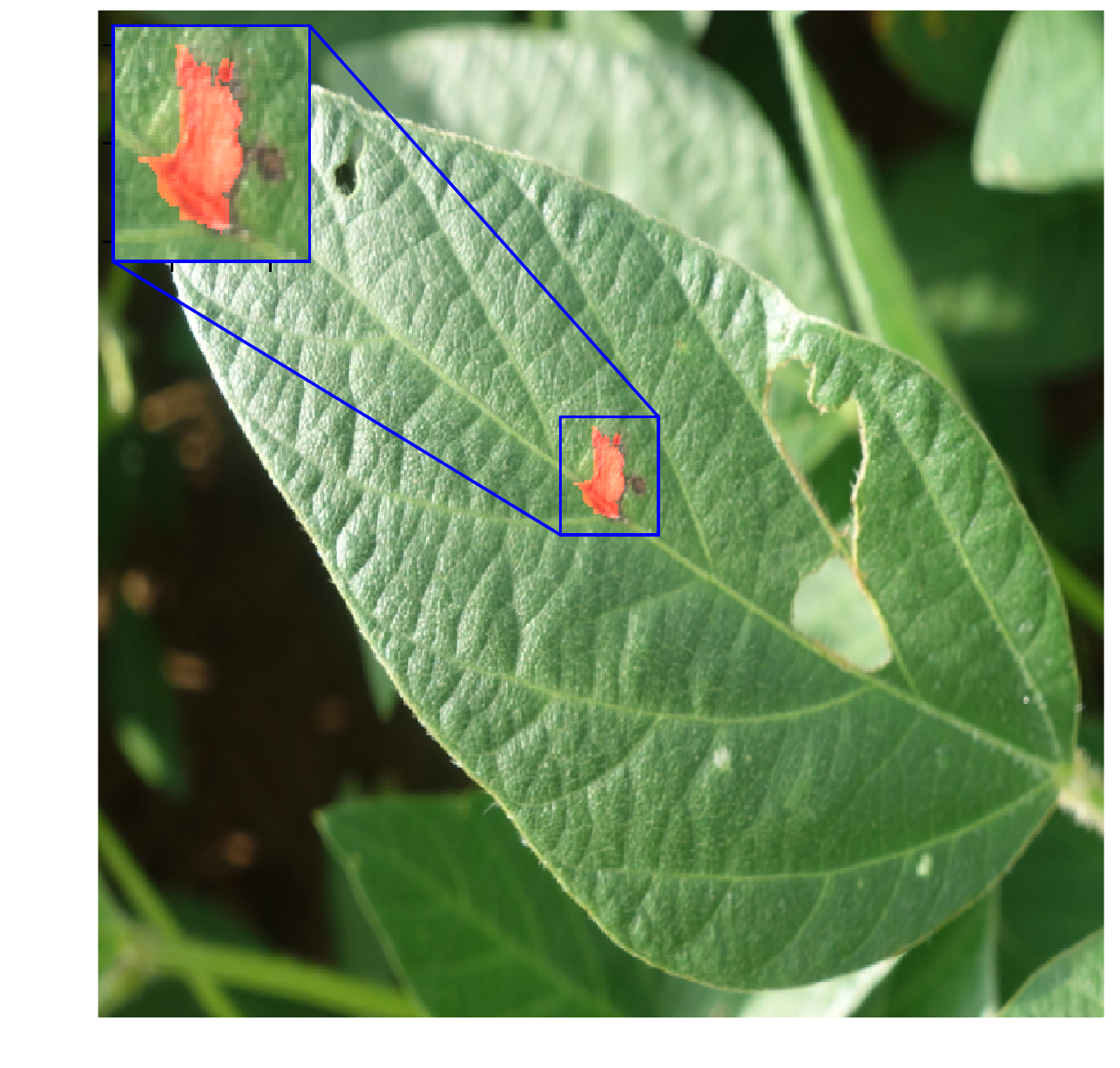}}
	\subfigure[FCN]{\includegraphics[width=.32\columnwidth]{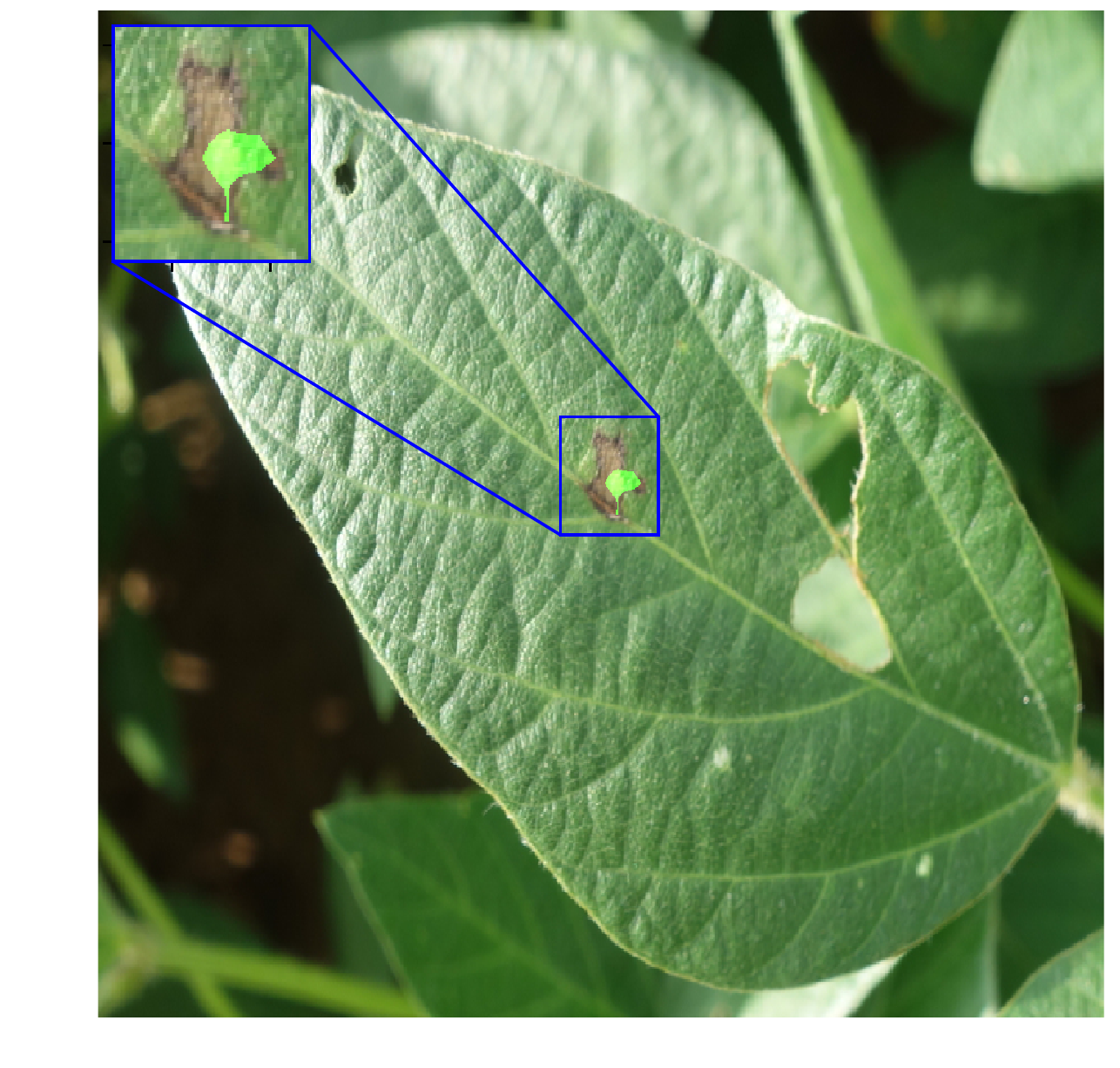}}
	\subfigure[Proposed Approach]{\includegraphics[width=.32\columnwidth]{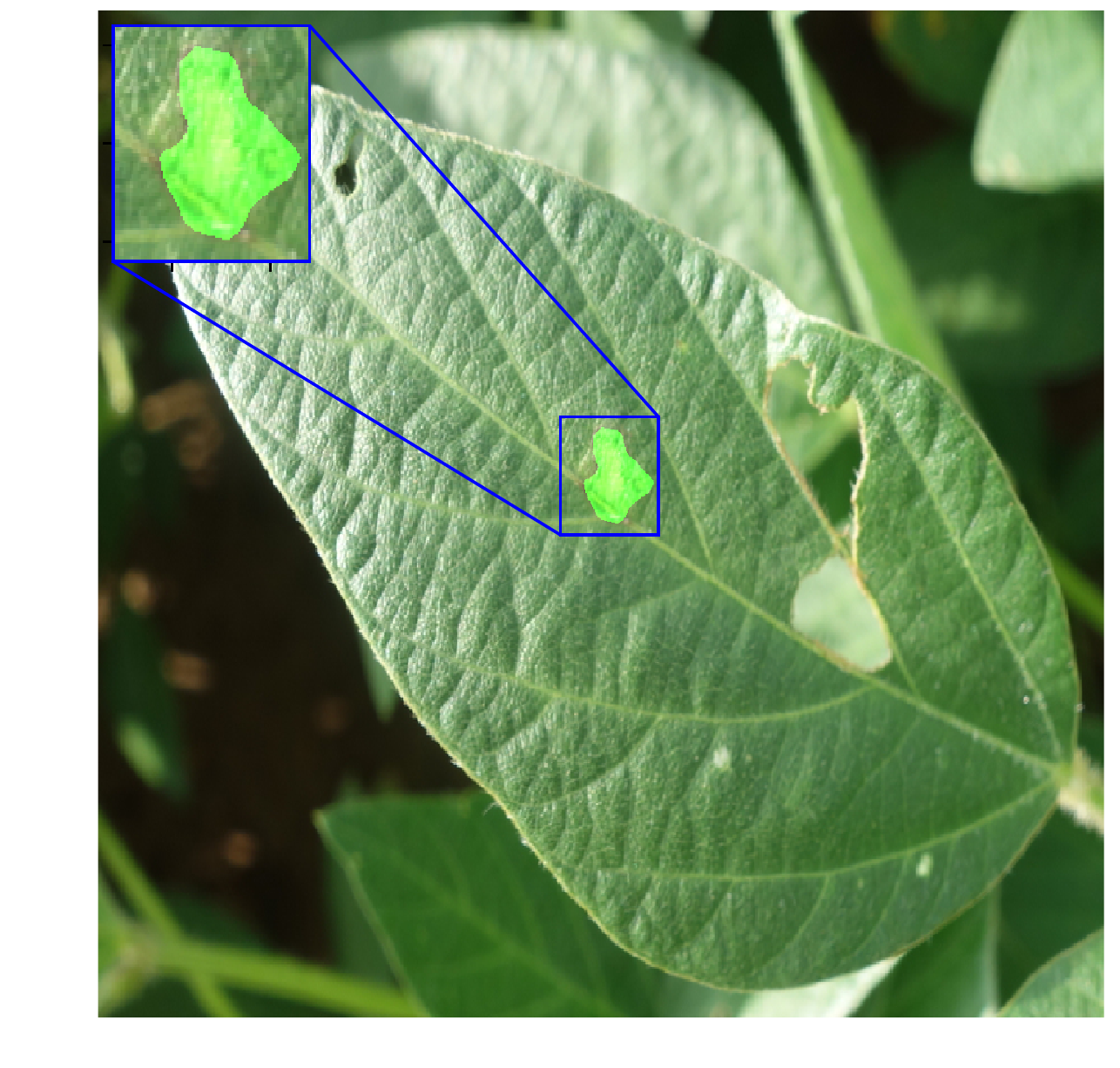}}
	
	\caption{\label{fig:soybean_dataset} Example of (a) ground-truth, (b) FCN and (c) proposed approach from Soybean Disease dataset.}
\end{figure}

\textbf{Soybean Disease Dataset.}
As shown in Fig. \ref{fig:soybean_dataset}, the proposed approach was able to segment soybean disease with excellent pixel accuracy. It detects regions of disease that the baseline was not capable of, as illustrated in the second example. The proposed approach also segments the disease pixels more accurately compared to the baseline (see the third example). However, the proposed approach generally segments a region larger than the ground-truth, which explains the lower IoU compared to the baseline. In this task, it is important to have a low false-negative (as in the proposed approach) to detect diseases early and reduce losses.

\subsection{Noise Invariance}

\begin{figure}[!ht]
	\centering
	\subfigure{\includegraphics[width=.4\columnwidth]{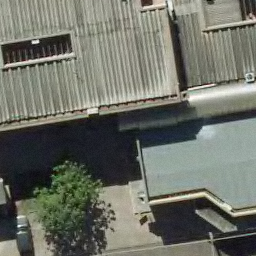}}
	\subfigure{\includegraphics[width=.4\columnwidth]{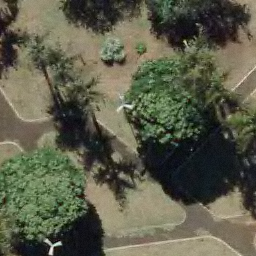}}
	\subfigure{\includegraphics[width=.4\columnwidth]{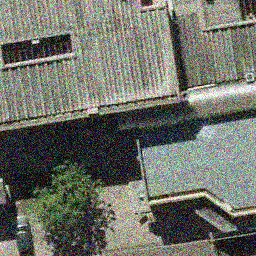}}
	\subfigure{\includegraphics[width=.4\columnwidth]{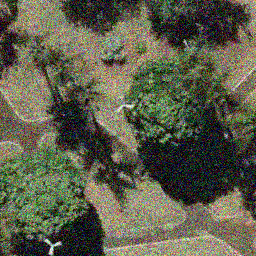}}
	
	\caption{\label{fig:noise_images2} Original images and their respective noisy images.}
\end{figure}

Noise invariance of semantic segmentation methods was assessed on the Urban Tree dataset.
Gaussian noise with $\sigma = 0.02$ was added to the images as illustrated in Fig. \ref{fig:noise_images2}.
We trained the proposed approach and the FCN baseline using noisy images.
Then, we evaluated them in the test set with and without noise.

The results using noisy images in the training of both approaches are shown in Table \ref{tab:results_noise}. The second column of the table presents the results using noisy test images. As expected, both approaches still provided good results as they were trained and tested on noisy images. Our approach has achieved superior pixel accuracy and IoU compared to the baseline (e.g., 0.875 versus 0.776 and 0.697 versus 0.686).

Although these results are promising, it is not possible to guarantee that the method discarded noise in training, since the test images were also noisy. To effectively assess the noise invariance, the third column of Table \ref{tab:results_noise} shows the results using noisy images in the training and noise-free images in the test. The baseline FCN presented weak results, showing that the noise had great interference in its training. On the other hand, the proposed approach showed consistent results, which demonstrates its robustness to noise. Our approach obtained pixel accuracies of 0.875 and 0.847 in test images with and without noise, a drop of only 0.028.

\begin{figure}[!ht]
	\centering
	\subfigure[Noisy test image]{\label{fig:noise_imagesa}\includegraphics[width=.32\columnwidth]{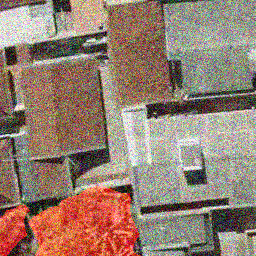}}
	\subfigure[FCN]{\label{fig:noise_imagesb}\includegraphics[width=.32\columnwidth]{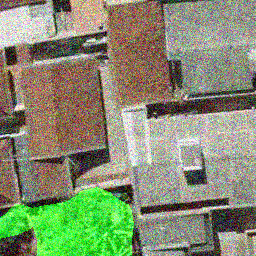}}
	\subfigure[Proposed Approach]{\label{fig:noise_imagesc}\includegraphics[width=.32\columnwidth]{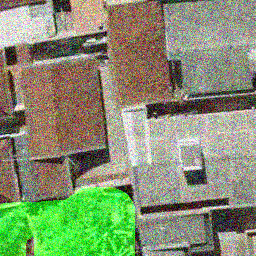}}
	\subfigure[Noisy-free test image]{\label{fig:noise_imagesd}\includegraphics[width=.32\columnwidth]{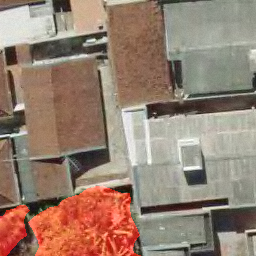}}
	\subfigure[FCN]{\label{fig:noise_imagese}\includegraphics[width=.32\columnwidth]{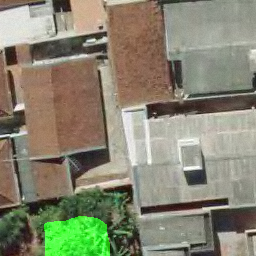}}
	\subfigure[Proposed Approach]{\label{fig:noise_imagesf}\includegraphics[width=.32\columnwidth]{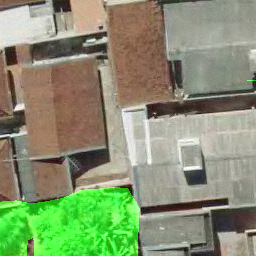}}
	
	\subfigure[Noisy-free test image]{\label{fig:noise_imagesg}\includegraphics[width=.32\columnwidth]{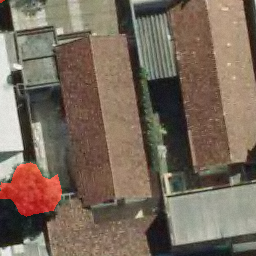}}
	\subfigure[FCN]{\label{fig:noise_imagesh}\includegraphics[width=.32\columnwidth]{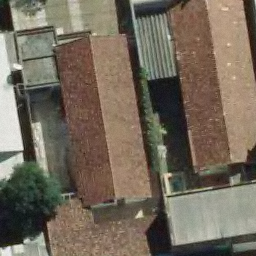}}
	\subfigure[Proposed Approach]{\label{fig:noise_imagesi}\includegraphics[width=.32\columnwidth]{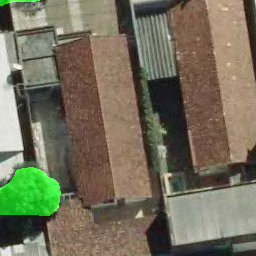}}
	
	\caption{\label{fig:noise_images} Comparative results of the proposed approach and FCN trained in noisy images. The first row of images shows the segmentation using a noisy test image while the second row of images shows the results using a noisy-free test image.}
\end{figure}

Fig. \ref{fig:noise_images} shows the visual segmentation results of both methods in test images with and without noise. The results of the baseline FCN and the proposed approach in a noisy test image (Fig. \ref{fig:noise_imagesa}) are shown in Figs. \ref{fig:noise_imagesb} and \ref{fig:noise_imagesc}, respectively. As the methods were trained on noisy images, they achieved satisfactory results despite the apparent noise. However, when a noisy-free image is used in testing methods trained with noisy images, the results of the proposed approach are superior to FCN as shown in Figs. \ref{fig:noise_imagesd}- \ref{fig:noise_imagesi}.

\begin{table}[!ht]
\begin{center}
\caption{Comparative results between our method and the baseline FCN using noisy images to train.}
\label{tab:results_noise}
\begin{tabular}{|c|c|c||c|c|}
\hline
\multicolumn{1}{|c|}{\multirow{2}{*}{\textbf{Method}}} & \multicolumn{2}{c||}{\textbf{Noisy Images}} & \multicolumn{2}{c|}{\textbf{Noise-free Images}} \\ \cline{2-5} 
\multicolumn{1}{|c|}{}                        & \textbf{PA}          & \textbf{IoU}         & \textbf{PA}           & \textbf{IoU} \\ \hline
FCN  & 0.776 & 0.686 & 0.122 & 0.122 \\ \hline
Ours & 0.875 & 0.697 & 0.847 & 0.569 \\ \hline
\end{tabular}
\end{center}
\end{table}

\section{Conclusion}

A correctly weighted loss is important for semantic segmentation methods, mainly in datasets with imbalanced classes and labeling uncertainty. This paper shows how these challenges can be considered in a new loss function. The proposed approach combines two weights: i) the importance of the class given its occurrence and ii) the uncertainty in the labeling of pixels close to the edges. To the best of our knowledge, this is the first approach that overcomes both challenges using pixel-wise weights during training.

The robustness of the proposed approach can be ascertained for the three datasets considered; which present different characteristics and challenges. The results showed that the proposed approach obtains superior metrics regardless of the segmentation method (e.g., SegNet and FCN). Significant results with an increase of up to 40\% in accuracy were achieved by the proposed approach, which clearly shows its relevance. Our approach also proved to be more invariant to noise, even when training was performed on noisy images and tested on noise-free images.

As a future work, we intend to evaluate new semantic segmentation methods. Further research also includes the application of the proposed approach to segmentation problems with several classes.

\section*{Acknowledgments.}
This study was supported by the FUNDECT - State of Mato Grosso do Sul Foundation to Support Education, Science and Technology, CAPES - Brazilian Federal Agency for Support and Evaluation of Graduate Education, and CNPq - National Council for Scientific and Technological Development. The Titan V and XP used for this research was donated by the NVIDIA Corporation.

\bibliographystyle{unsrtnat}


\end{document}